\documentclass{article}
\usepackage{ijcai26}

\usepackage{times}
\usepackage{soul}
\usepackage{url}
\usepackage[hidelinks]{hyperref}
\usepackage[utf8]{inputenc}
\usepackage[small]{caption}
\usepackage{graphicx}
\usepackage{amsmath}
\usepackage{amsthm}
\usepackage{booktabs}
\usepackage{algorithm}
\usepackage{algorithmic}
\usepackage{subcaption}
\usepackage[switch]{lineno}
\usepackage{tikz}
\usetikzlibrary{shapes.geometric, arrows.meta, positioning, shadows, fit}
\usepackage{pgfplots}
\usepackage{fontawesome5}
\usepackage{listings}
\pgfplotsset{compat=1.18}
\usepgfplotslibrary{statistics}

\title{Physics-in-the-Loop: A Hybrid Agentic Architecture for Validated CAD Engineering Design}

\author{
Elias Berger$^{1,2}$
\And
Muhammad Usama$^{3,4}$\And
Jan Mehlstäubl$^{2}$\And
Bernhard Saske$^1$\And
Kristin Paetzold-Byhain$^1$\\
\affiliations
$^1$Dresden University of Technology, Dresden, Germany\\
$^2$MAN Truck \& Bus SE, Munich, Germany\\
$^3$German Research Center for Artificial Intelligence, Kaiserslautern, Germany\\
$^4$RPTU Kaiserslautern-Landau, Germany\\
\emails
\{elias.berger, bernhard.saske, kristin.paetzold-byhain\}@tu-dresden.com,
jan.mehlstaeubl@digitalhub.man,
wop76ziga@dfki.de
}

\begin{document}

\maketitle

\begin{abstract}
    Large Language Models (LLMs) can generate Computer-Aided Design (CAD), yet lack physical comprehension required for reliable engineering design. Instead of attempting to implicitly learn physical laws from data, we propose a Hybrid Agentic-Physical Architecture that embeds validated knowledge-based engineering tools directly into the decision-making loop of autonomous AI agents. In this framework, engineering design is formulated as a closed-loop, sequential decision-making process guided by explicit physical verification. Based on a load case, dedicated agents iteratively plan, generate, evaluate, and revise engineering designs using knowledge-based tools as a feedback signal. We introduce a benchmark dataset and metrics for assessing functional validity in generative CAD. Our system generates more complex and physically verified designs, with a 4.2$\times$ increase in structural complexity and improving compile rate by 3.5\% compared to similar agentic methods. The codebase, prompts and dataset will be made publicly available to support reproducibility and future research. 

\end{abstract}

\section{Introduction}

\begin{figure}[t]
    \centering
    \begin{minipage}{0.32\linewidth}
        \includegraphics[width=0.9\linewidth]{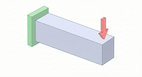} \\
        \includegraphics[width=\linewidth]{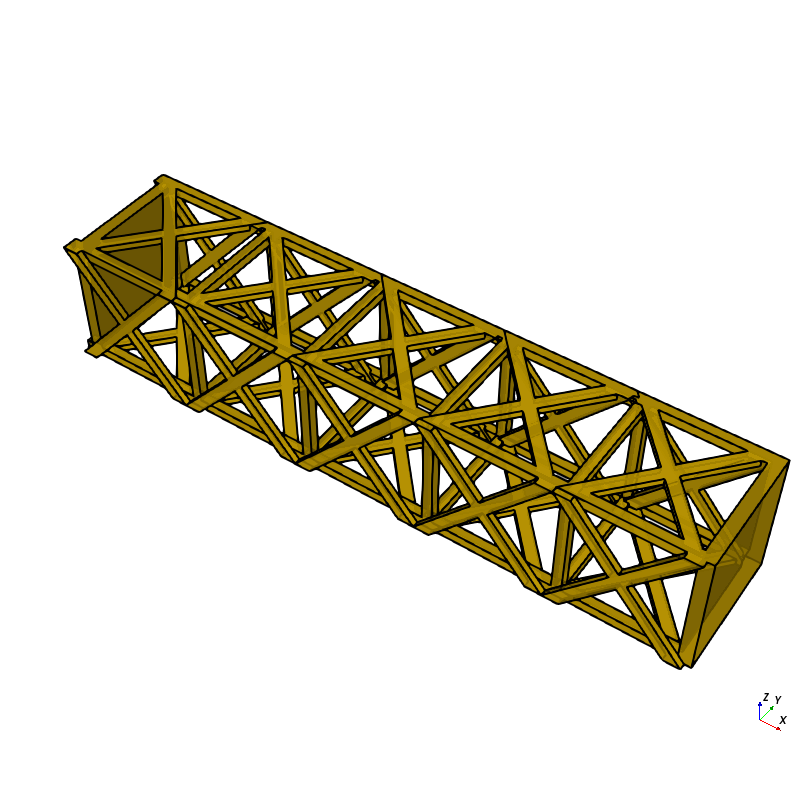}
    \end{minipage}
    \hfill
    \begin{minipage}{0.32\linewidth}
        \includegraphics[width=0.8\linewidth]{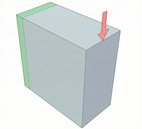} \\
        \includegraphics[width=\linewidth]{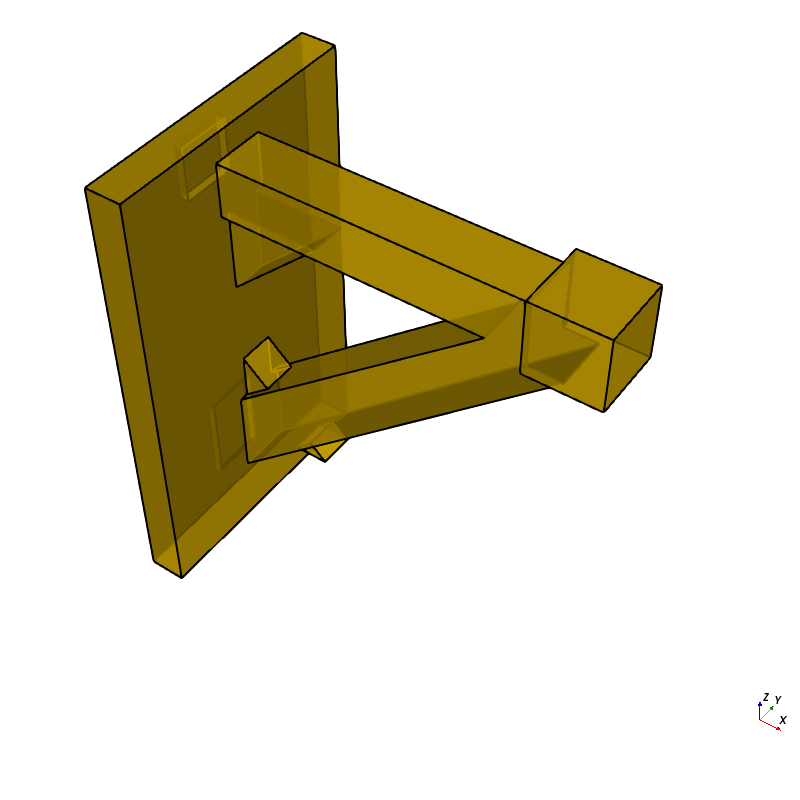}
    \end{minipage}
    \hfill
    \begin{minipage}{0.32\linewidth}
        \includegraphics[width=\linewidth]{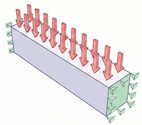} \\
        \includegraphics[width=\linewidth]{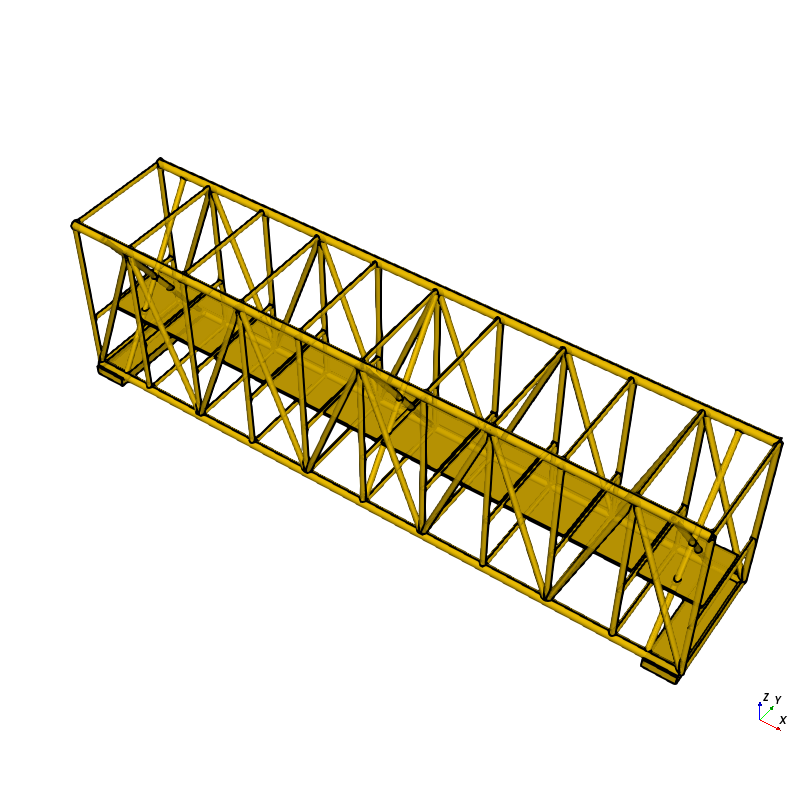}
    \end{minipage}
    \caption{Generated CAD design examples. Top row: Input load cases defining design space, boundary conditions and forces. Bottom row: Resulting geometries satisfying the physical constraints.}
    \label{fig:intro_teaser}
\end{figure}
Generative artificial intelligence (AI) has recently gained momentum in engineering design  \cite{Zhang2025,steiningerEnhancingComputeraidedDesign2025,REGASSAHUNDE2022100478}. Notwithstanding this progress, transferring these methods into real-world engineering practice, where reliable and validated structural integrity is essential, remains largely an open challenge. \cite{berger2025challenges,preintner2025evocadevolutionarycadcode}.

LLM-based generative methods enable the synthesis of parametric CAD models from textual and multimodal inputs \cite{xuCADMLLMUnifyingMultimodalityConditioned2025}. However, existing generative CAD systems are predominantly assessed using geometric similarity metrics, which fail to capture functional validity, load-bearing capacity, or other engineering objectives \cite{Berger2026FromGeometryToFunction}. Generated designs appear visually plausible but lack guarantees of functional correctness or technical feasibility. Recent efforts have attempted to improve reliability and complexity through iterative feedback using Vision Language Models (VLM) to visually verify 2D renders of CAD outputs \cite{alrashedyGeneratingCADCode2025,preintner2025evocadevolutionarycadcode}. While effective for shape fidelity, visually-evaluated systems do not verify the structural integrity of the generated parts and produce very simple geometries. These approaches use the DeepCAD dataset \cite{wuDeepCADDeepGenerative2021}, which contains mostly very simple geometries with no labels relevant for engineering design.

In this paper, we formulate generative CAD as a physics-constrained engineering problem. We propose an Agentic CAD System that uses VLMs to automate mechanical design through an iterative ``Generate-Simulate-Refine'' loop. Unlike previous works that rely on visual feedback, our method integrates a validated knowledge-based tool into the feedback loop. Our agents receive feedback from physics-based tools to iteratively enhance the output to be not only geometrically valid but structurally sound. We do not train VLMs, but make use of the multi-task in-context capabilities of VLMs. Thus our approach does not rely on large annotated datasets. Further, we define new metrics and publicize a benchmark dataset for evaluating functional validity in generative CAD. Our method does not depend on existing training datasets, as we leverage the multi-task capabilities of LLMs and VLMs. This allows us to generate designs of high complexity, exceeding prior works by a factor of three in geometric complexity (examples see Figure \ref{fig:intro_teaser}).

Our contributions are as follows:
\begin{enumerate}
    \item A hybrid multi-agent architecture integrating physics-in-the-loop active decision signal that generates CAD design with exceeding complexity.
    \item Systematic comparison of generative, agentic, and hybrid AI paradigms and empirical evidence that physics-guided agents improve reliability and physical validity.
    \item A novel benchmark for functional CAD generation under load bearing requirements. We make the load case data available at camera-ready, including the dataset, evaluation scripts, and agent prompt templates using MIT and CC licenses.
\end{enumerate}


\section{Related Work}

\subsection{LLMs for CAD and Engineering Design}

DeepCAD \cite{wuDeepCADDeepGenerative2021} is the initial paper that pioneered deep learning methods for CAD generation using a design history. Since then several studies demonstrate the ability of LLMs to generate parametric CAD models from textual \cite{khanText2CADGeneratingSequential2024,wangCADGPTSynthesisingCAD2025,LvBao2025CADInstruct,wangTexttoCADGenerationInfusing2025,govindarajanCADmiumFineTuningCode2026,usama2025nurbgenhighfidelitytexttocadgeneration,Berger2026MultiTaskCG}, image \cite{chenImg2CADConditioned3D2025,alamGenCADImageConditionedComputerAided2025}, point cloud \cite{dupont2024transcadhierarchicaltransformercad,khanCADSIGNetCADLanguage2024}, or multi-modal \cite{dorisCADCoderOpenSourceVisionLanguage2025,xuCADMLLMUnifyingMultimodalityConditioned2025,kolodiazhnyiCadrilleMultimodalCAD2025} inputs.

Agentic and iterative generation processes have been recently proposed. \citeauthor{preintner2025evocadevolutionarycadcode} integrates an evolutionary optimization loop to iteratively improve generated designs based on visual feedback. \citeauthor{ockerIdeaCADLanguage2025} propose a Multi-Agent System driven by a VLM to automate CAD model generation by mirroring the structure of human engineering teams with specialized agents. \citeauthor{alrashedyGeneratingCADCode2025} introduces CADCodeVerify, a method that employs commercial VLMs and prompting to generate CAD code and uses visual tests to verify if the CAD object matches the intended shape.

A shared challenge of these methods is the reliance on a limited training dataset. Public datasets such as DeepCAD \cite{wuDeepCADDeepGenerative2021}, Fusion360 \cite{willisFusion360Gallery2021}, and CADParser \cite{zhouCADParserLearningApproach2023} contain only a few ten thousand CAD models with little complexity, which is small compared to datasets in other domains.

Other approaches operate on different representations such as Boundary Representation (B-Rep) \cite{jayaraman2023solidgenautoregressivemodeldirect,lambourne2021brepnettopologicalmessagepassing} or meshes \cite{nashPolyGenAutoregressiveGenerative2020} but do not produce  CAD models with a design history.

Further, the evaluation of these generative CAD methods has so far been primarily based on geometric similarity metrics such as Chamfer Distance, Intersection-over-Union (IoU), or Normal Consistency. While optimizing replication capabilities, the metrics lack consideration of real-world functional requirements like load-bearing capacity \cite{preintner2025evocadevolutionarycadcode,heCADCoderTextGuidedCAD2025}. Consequently, this research area would benefit from new data and metrics.

\subsection{Agentic AI and Multi-Agent Systems}

Agentic AI commonly describes systems in which one or more LLMs are embedded within an execution loop. Agentic systems allow for task distribution, observation of intermediate outcomes, and iterative refinement towards a specified objective. \cite{plaat2025agenticllmsurvey,huang2024planningagents}. In contrast to single-turn text generation, agentic architectures tightly integrate reasoning with interaction in an external environment. This enables using feedback to improve task completion and reduce hallucinations without requiring updates to model parameters. \cite{yao2023react,shen2024llmwithtools}.

\subsection{Knowledge-based Engineering and Hybrid AI}

Knowledge-based engineering tools are explicit physical models and formalized design knowledge that, compared to LLMs, have already reached a very mature state \cite{Herrmann2021}. Established tools such as topology optimization and FEA use physics-based objectives and constraints to reliably produce functionally optimized designs, albeit within restricted domains and no learning on past data. The fundamental gap between generative creativity and validated engineering tools motivates hybrid approaches that combine data-driven AI with knowledge-based tools.

\section{Problem Formulation}

With this work we aim to contribute towards AI for engineering design -- a discipline that focuses on the creation of technical parts that fulfill specific functional requirements \cite{Pahl2007}. Engineering design commonly follows an iterative process, where software tools such as CAD and FEA are used to design and validate artifacts digitally \cite{Hirz2011}. Besides the geometric appearance of an artifact in engineering design its structural and functional performance under physical loads is of key importance \cite{schulteFunctionalFeaturesDesign1993a}. The input for the problem is a structured load case defining how a part is constrained at fixed supports, the forces acting upon it, and the design space within which it is defined. The task is to generate a CAD model that satisfies the load case.

We use an established PyTorch-based FEA solver\footnote{https://github.com/meyer-nils/torch-fem} to evaluate the structural performance and the objective of the AI system is to achieve a safety factor between 2.0 and 5.0, consistent with engineering practice \cite{budynas2020shigley} while minimizing the volume of the part. We use gmsh\footnote{https://gmsh.info} for meshing with optimization enabled for robustness. We set the material properties to match Aluminum 8081.

\section{Load Case Data}

\begin{figure*}
\centering
\setlength{\tabcolsep}{1pt}
\renewcommand{\arraystretch}{0.5}
\resizebox{0.88\linewidth}{!}{%
\begin{tabular}{c cccccccccc}
\rotatebox{90}{\scriptsize Load Case} 
 & \includegraphics[width=0.11\linewidth]{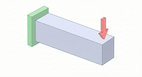}
 & \includegraphics[width=0.09\linewidth]{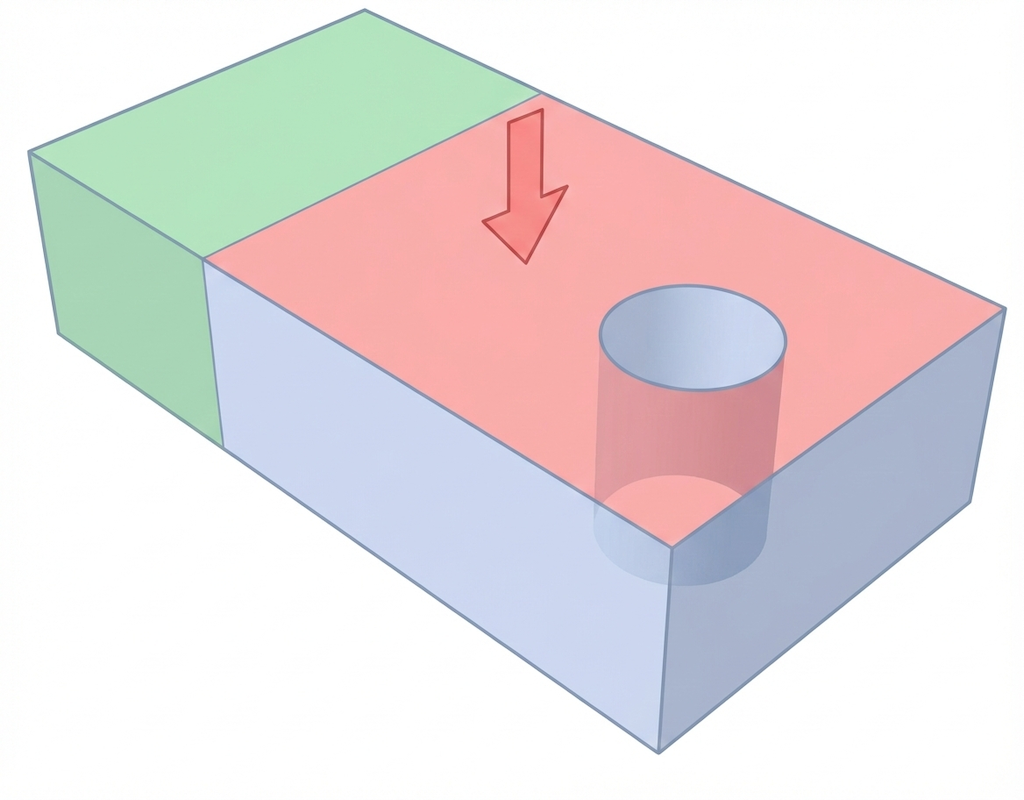}
 & \includegraphics[width=0.05\linewidth]{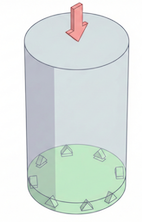}
 & \includegraphics[width=0.10\linewidth]{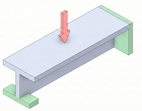}
 & \includegraphics[width=0.09\linewidth]{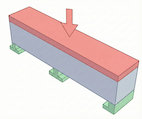}
 & \includegraphics[width=0.09\linewidth]{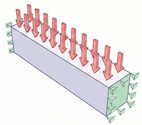}
 & \includegraphics[width=0.08\linewidth]{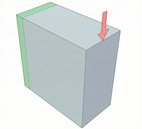}
 & \includegraphics[width=0.08\linewidth]{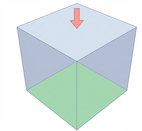}
 & \includegraphics[width=0.09\linewidth]{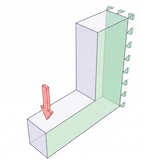} 
 & \includegraphics[width=0.09\linewidth]{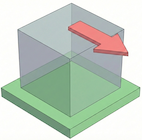}
 \\
 
 \rotatebox{90}{\scriptsize Gemini 3 Flash} &
 \includegraphics[width=0.09\linewidth]{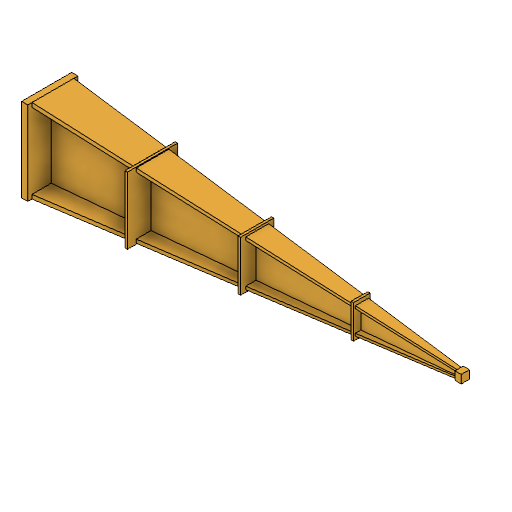} &
 \includegraphics[width=0.09\linewidth]{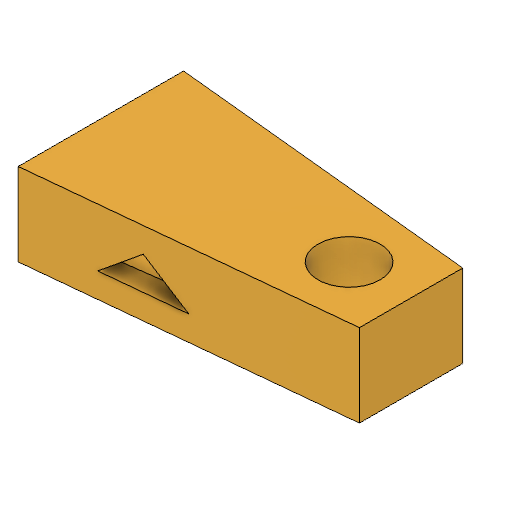} &
 \includegraphics[width=0.09\linewidth]{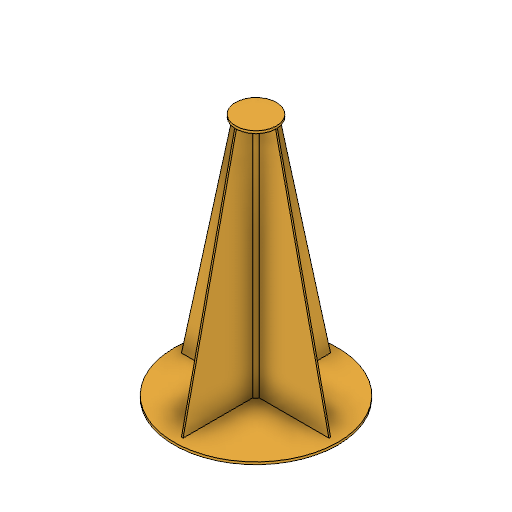} &
 \includegraphics[width=0.09\linewidth]{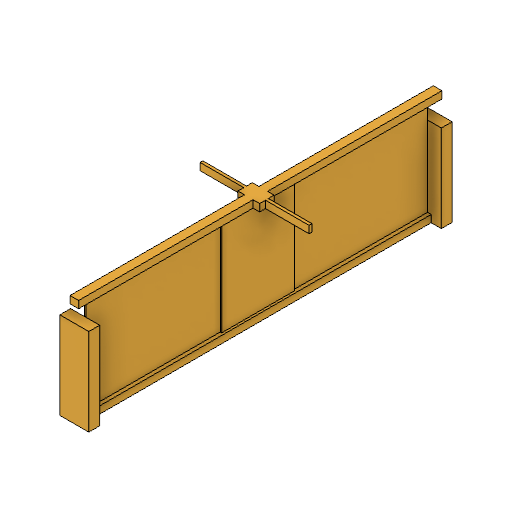} &
 \includegraphics[width=0.09\linewidth]{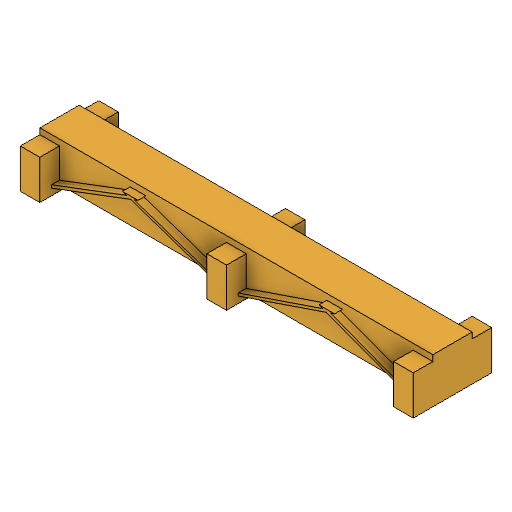} &
 \includegraphics[width=0.1\linewidth]{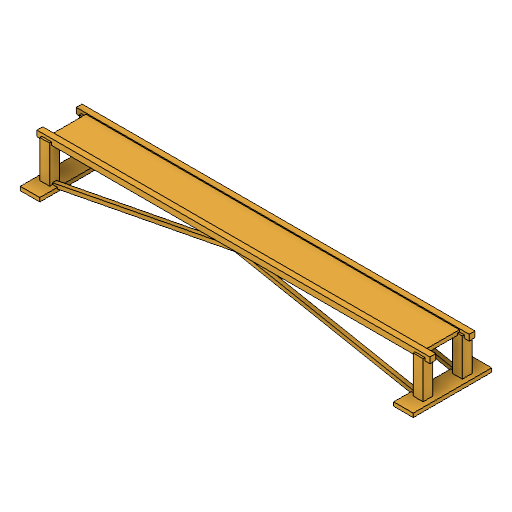} &
 \includegraphics[width=0.09\linewidth]{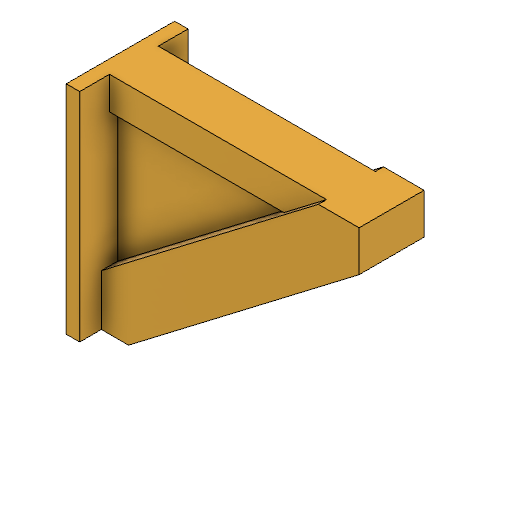} &
 \includegraphics[width=0.09\linewidth]{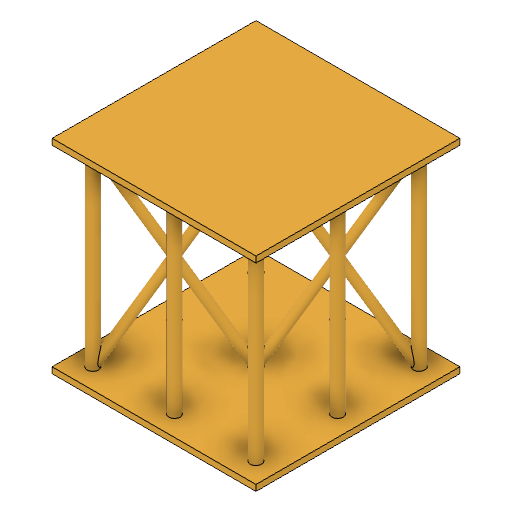} &
 \includegraphics[width=0.09\linewidth]{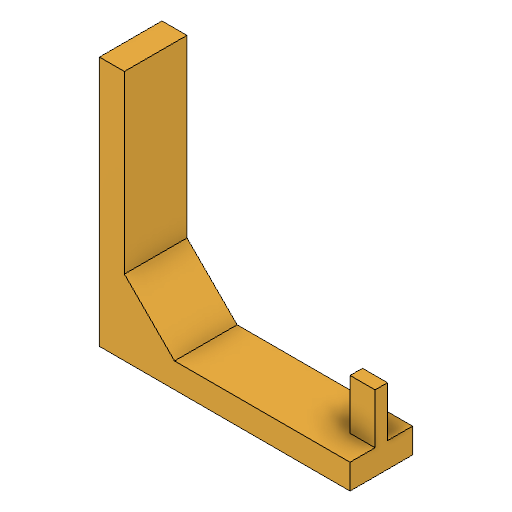} &
  \includegraphics[width=0.09\linewidth]{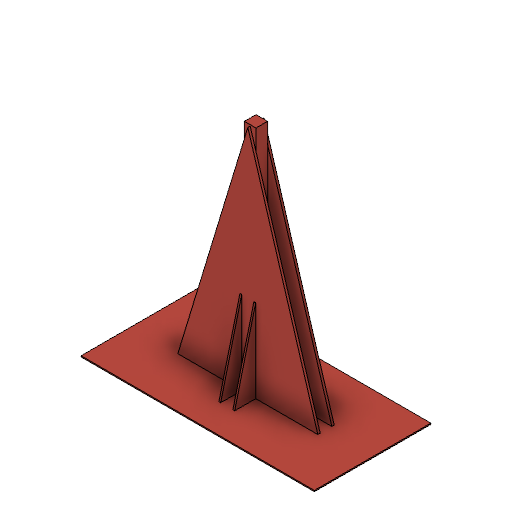}
 \\
 
 \rotatebox{90}{\scriptsize Gemini 3 Pro} &
 \includegraphics[width=0.09\linewidth]{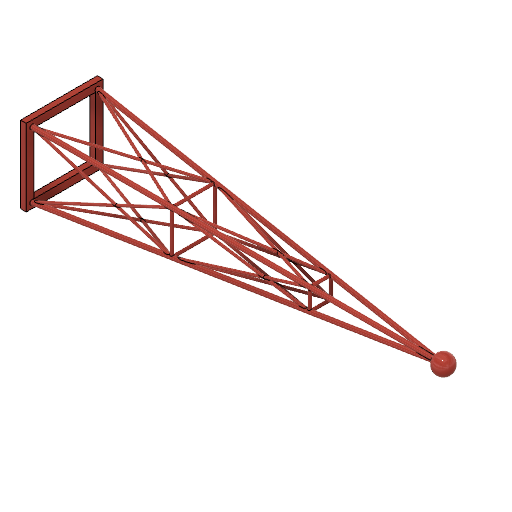} &
 \includegraphics[width=0.09\linewidth]{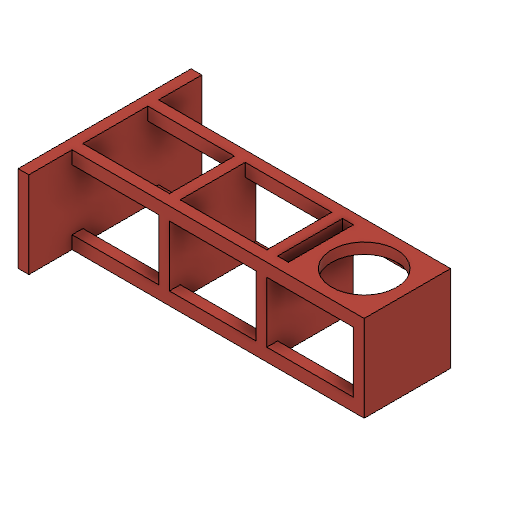} &
 \includegraphics[width=0.1\linewidth]{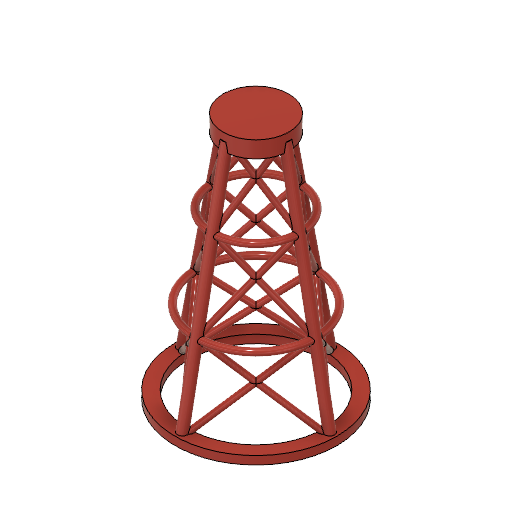} &
 \includegraphics[width=0.09\linewidth]{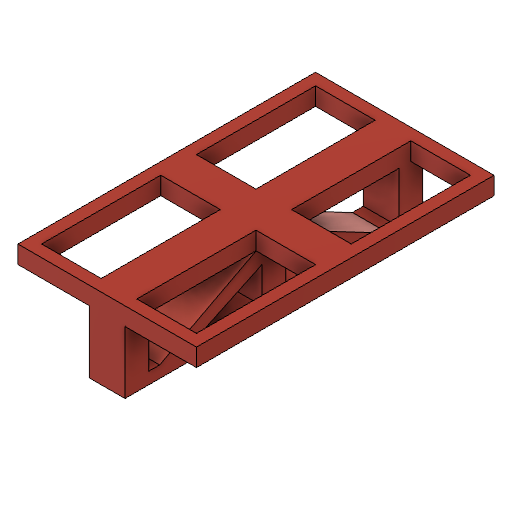} &
 \includegraphics[width=0.09\linewidth]{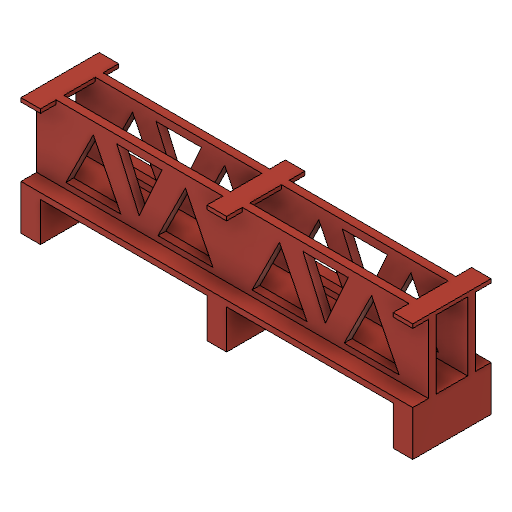} &
 \includegraphics[width=0.09\linewidth]{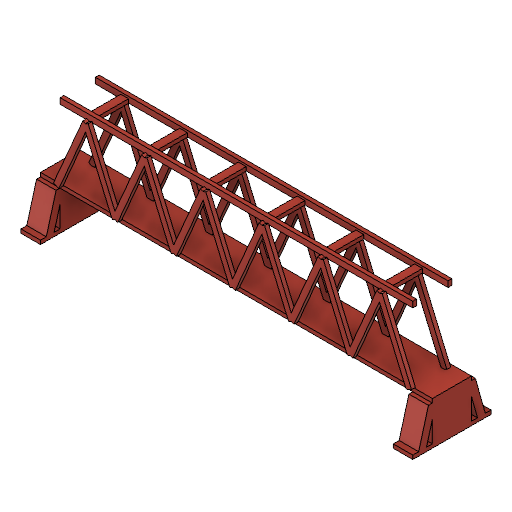} &
 \includegraphics[width=0.09\linewidth]{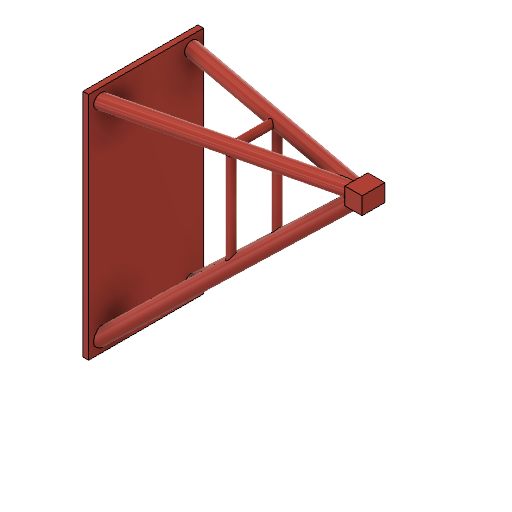} &
 \includegraphics[width=0.09\linewidth]{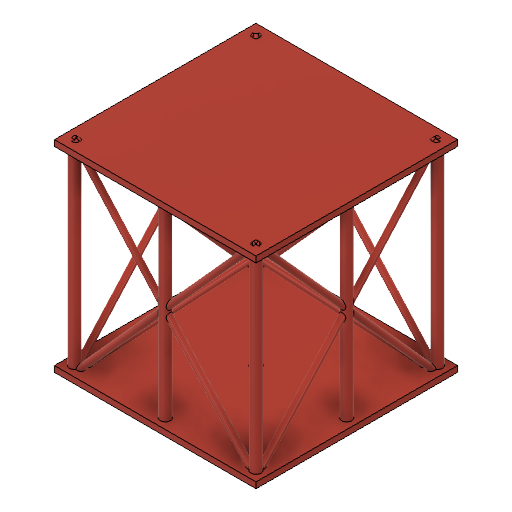} &
 \includegraphics[width=0.09\linewidth]{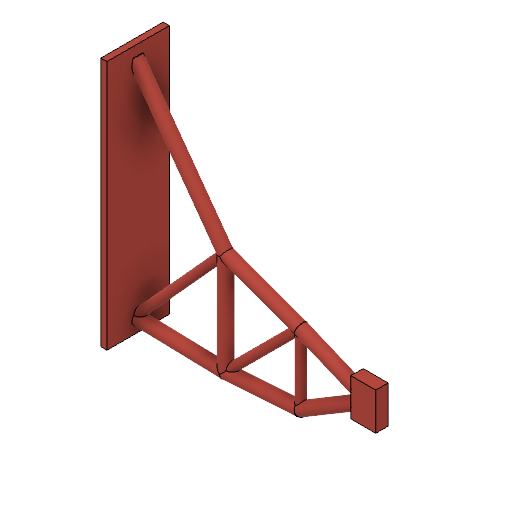} &
 \includegraphics[width=0.09\linewidth]{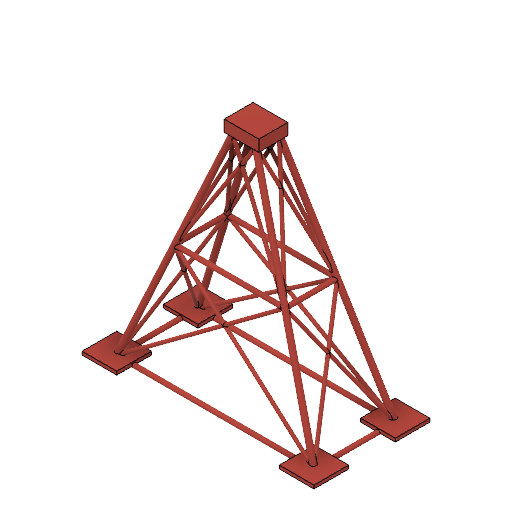} \\

 \rotatebox{90}{\scriptsize Claude Sonnet 4.5} &
 \includegraphics[width=0.09\linewidth]{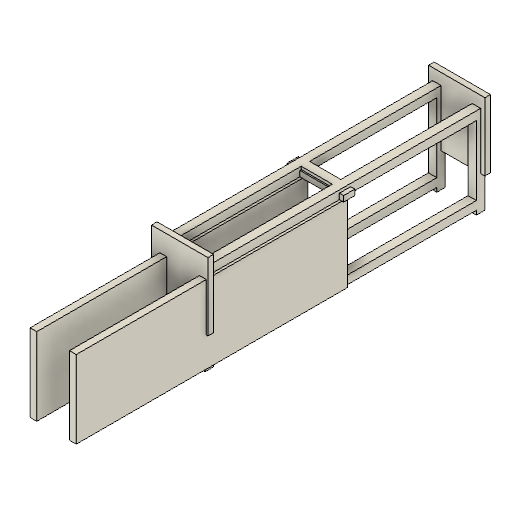} &
 \includegraphics[width=0.09\linewidth]{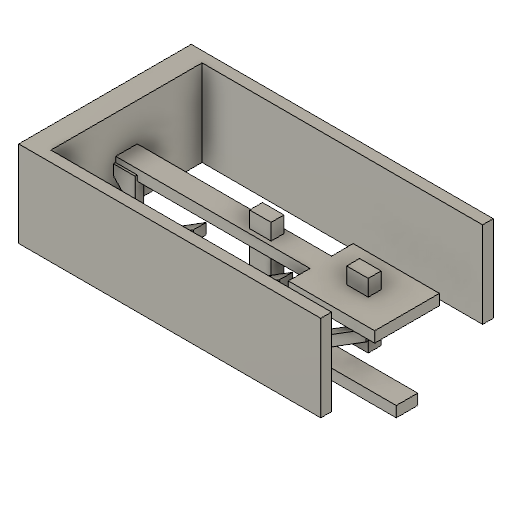} &
 \includegraphics[width=0.1\linewidth]{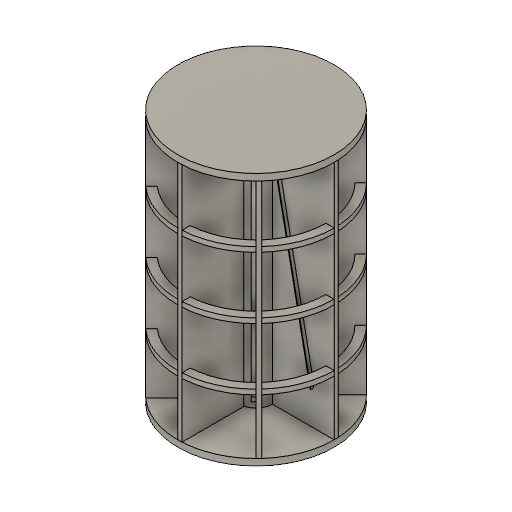} &
 \includegraphics[width=0.09\linewidth]{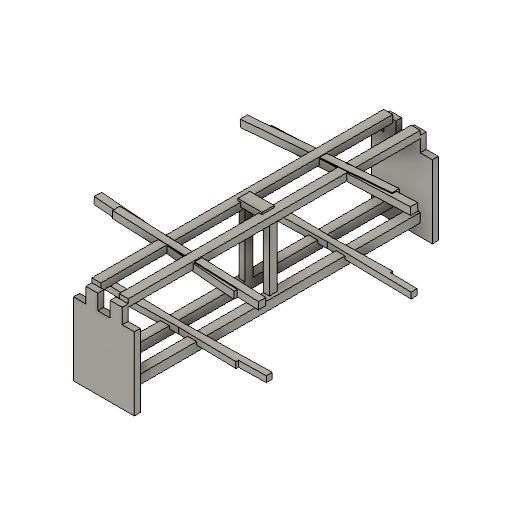} &
 \includegraphics[width=0.09\linewidth]{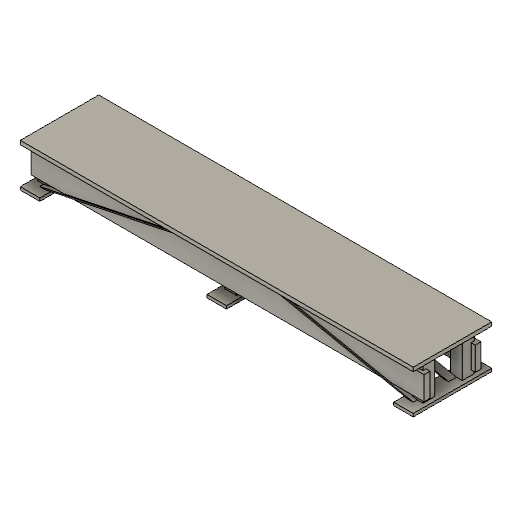} &
 \includegraphics[width=0.09\linewidth]{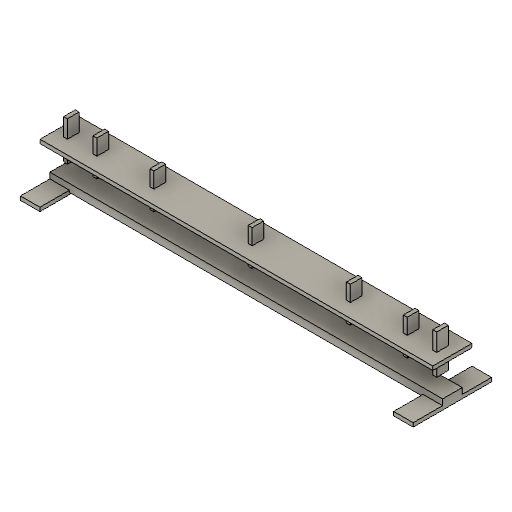} &
 \includegraphics[width=0.1\linewidth]{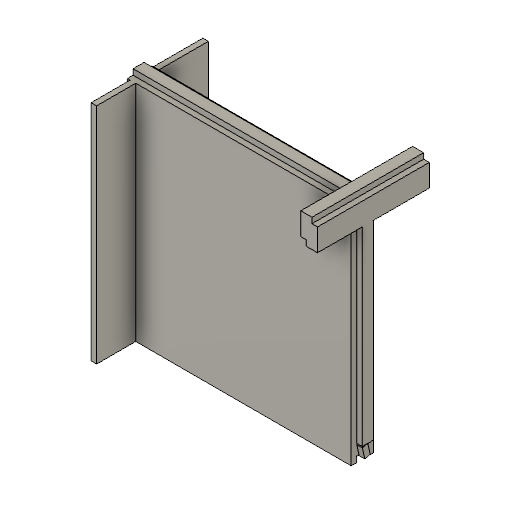} &
 \includegraphics[width=0.09\linewidth]{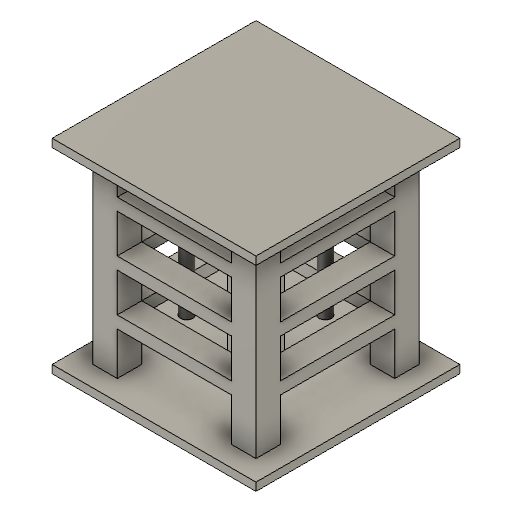} &
 \includegraphics[width=0.09\linewidth]{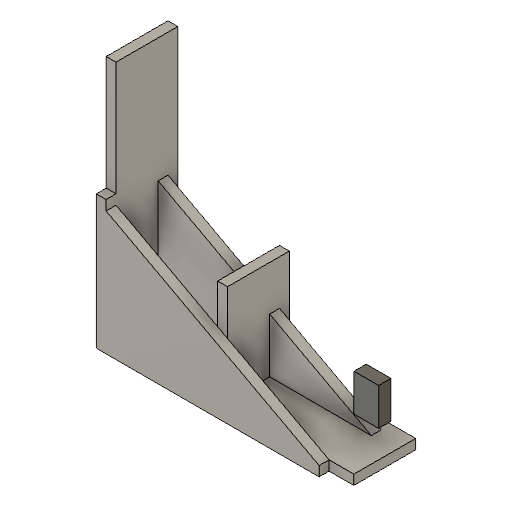} &
 \includegraphics[width=0.09\linewidth]{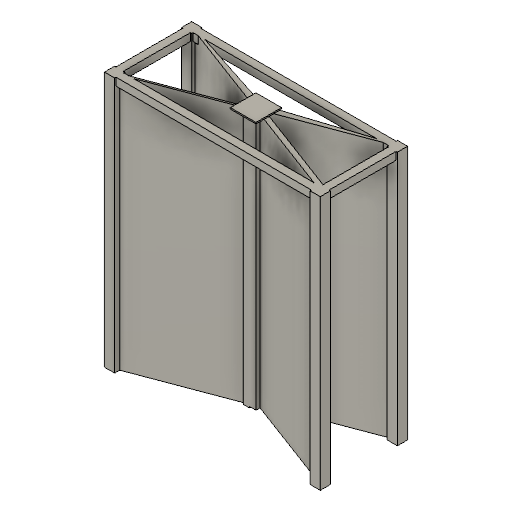} \\

 \rotatebox{90}{\scriptsize Claude Opus 4.5} &
 \includegraphics[width=0.09\linewidth]{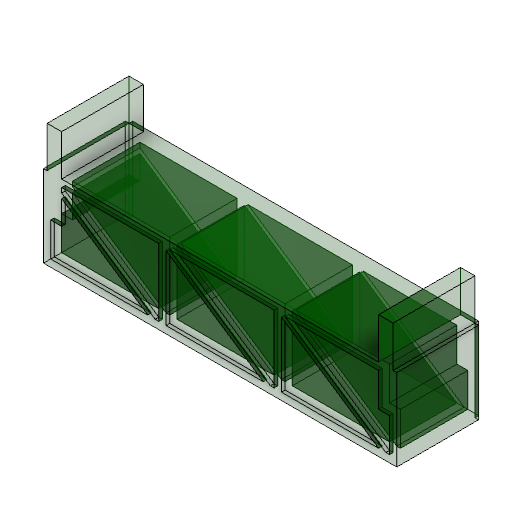} &
 \includegraphics[width=0.09\linewidth]{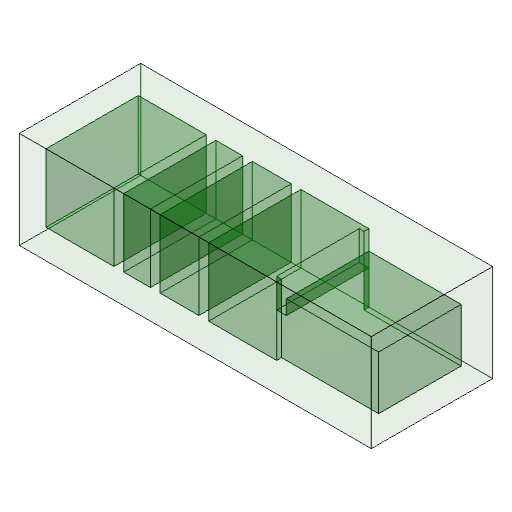} &
 \includegraphics[width=0.09\linewidth]{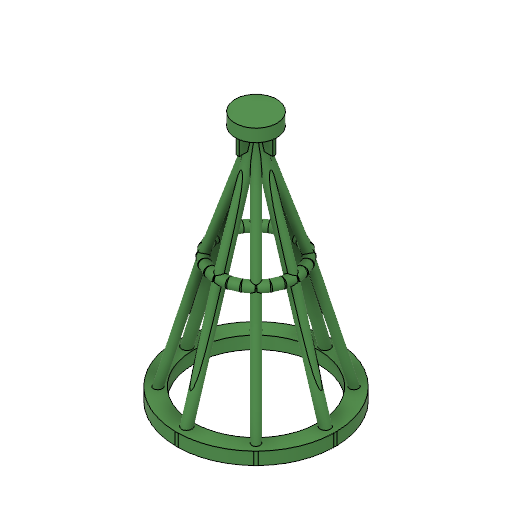} &
 \includegraphics[width=0.09\linewidth]{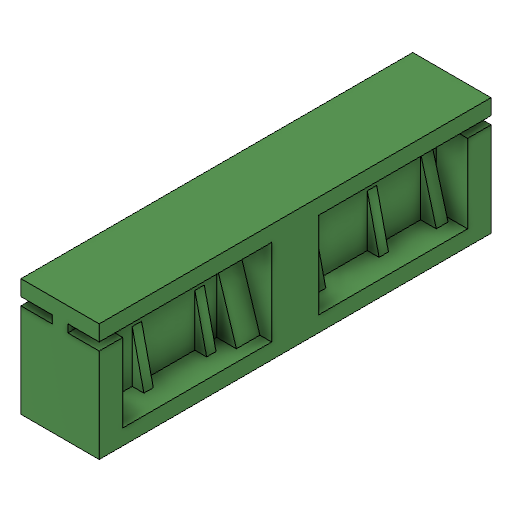} &
 \includegraphics[width=0.09\linewidth]{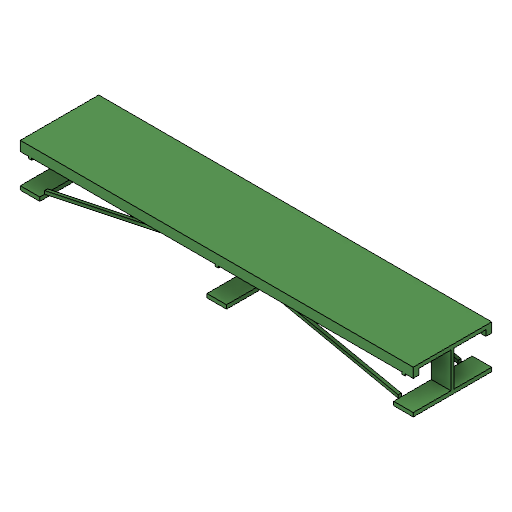} &
 \includegraphics[width=0.1\linewidth]{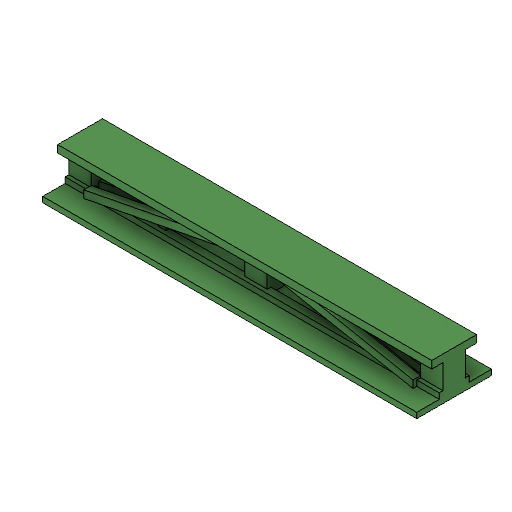} &
 \includegraphics[width=0.09\linewidth]{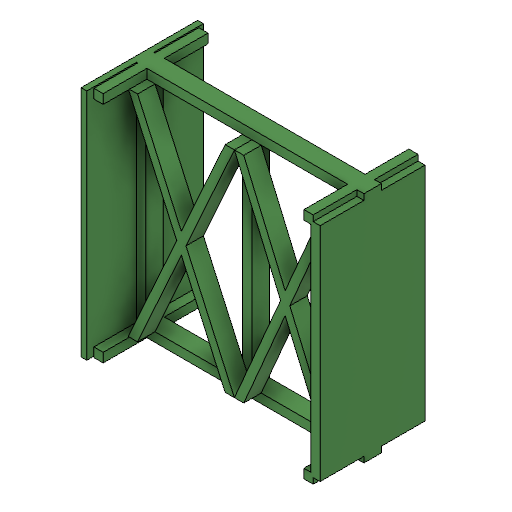} &
 \includegraphics[width=0.09\linewidth]{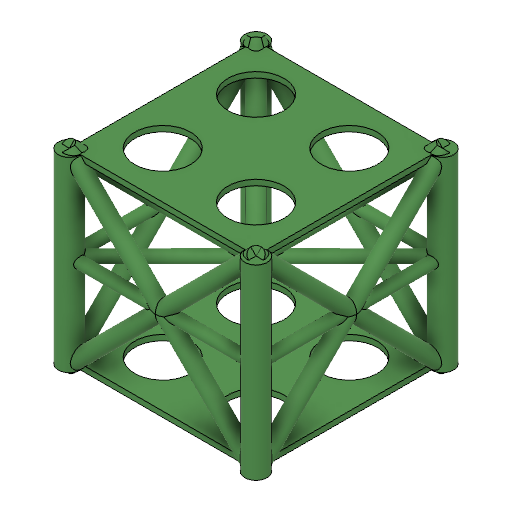} &
 \includegraphics[width=0.09\linewidth]{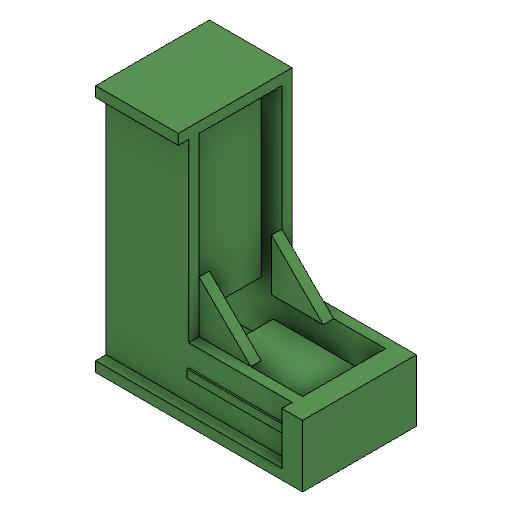} &
 \includegraphics[width=0.09\linewidth]{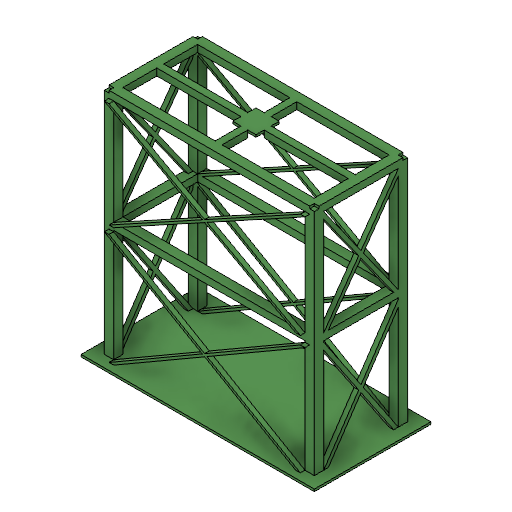} \\
\end{tabular}%
}
\caption{Generated CAD objects of the Hybrid Agentic Architecture. The top row shows visualizations of the input load case (red: forces, green: fixed supports, gray: design space). Subsequent rows show results from Gemini 3 Flash, Gemini 3 Pro, Claude Sonnet 4.5 , and Claude Opus 4.5.}
\label{fig:results}
\end{figure*}

We formulate the engineering design problem in terms of load bearing requirements that we formalize in load cases. Existing CAD datasets are not sufficient as they are not annotated with load case information necessary for physics-based validation. To this end, we construct a novel benchmark dataset comprising 20 representative load cases derived from standard mechanical design tasks \cite{Gaylord1990StructuralEngineeringHandbook}. Together with mechanical engineers we crafted the load cases to cover common challenges such as non-concave design spaces and internal holes. Each load case specifies fixed supports, applied forces, and explicit design space constraints. To evaluate generalization under varying constraints, we modify each load case using five distinct geometric scales and five force magnitude scales. As structural stiffness scales non-linearly with geometric size, variations in scale and loading induce distinct mechanical problems. In total, the benchmark comprises 500 unique load case configurations. Representative samples are illustrated in the left column of Figure~\ref{fig:results}. An example of the JSON schema is shown in Figure~\ref{fig:json_example}. We make the load case data available as JSON files and the CAD models as Python code. 

\begin{figure}[h]
\begin{lstlisting}[basicstyle=\ttfamily\footnotesize,breaklines=true,breakatwhitespace=false,columns=fullflexible]
{
  "meta": { "problem_id": "ARCH_BRIDGE", ... },
  "design_domain": {
    "units": "mm",
    "bounds": { "x_max": 1000.0, ... }
  },
  "spatial_selectors": [
    { "id": "support_left", "query": {
        "x_min": 0.0, "x_max": 50.0, ...
    } },
    ...
  ],
  "boundary_conditions": [
    { "spatial_selector_id": "support_left",
      "type": "fixed_displacement",
      "dof_lock": { ... } }
  ],
  "loads": [
    { "spatial_selector_id": "deck_surface",
      "type": "distributed_force",
      "magnitude_newtons": ...,
      "direction": ... },
  ]
}
\end{lstlisting}
\caption{Condensed excerpt of a load case JSON definition. \texttt{design\_domain} specifies the design space. \texttt{spatial\_selectors} define regions of interest for applying \texttt{boundary\_conditions} and \texttt{loads} to the selected regions. \texttt{dof\_lock} refers to degrees of freedom locked at fixed~supports.}
\label{fig:json_example}
\end{figure}


\begin{figure*}[t]
\centering
\includegraphics[width=0.95\linewidth]{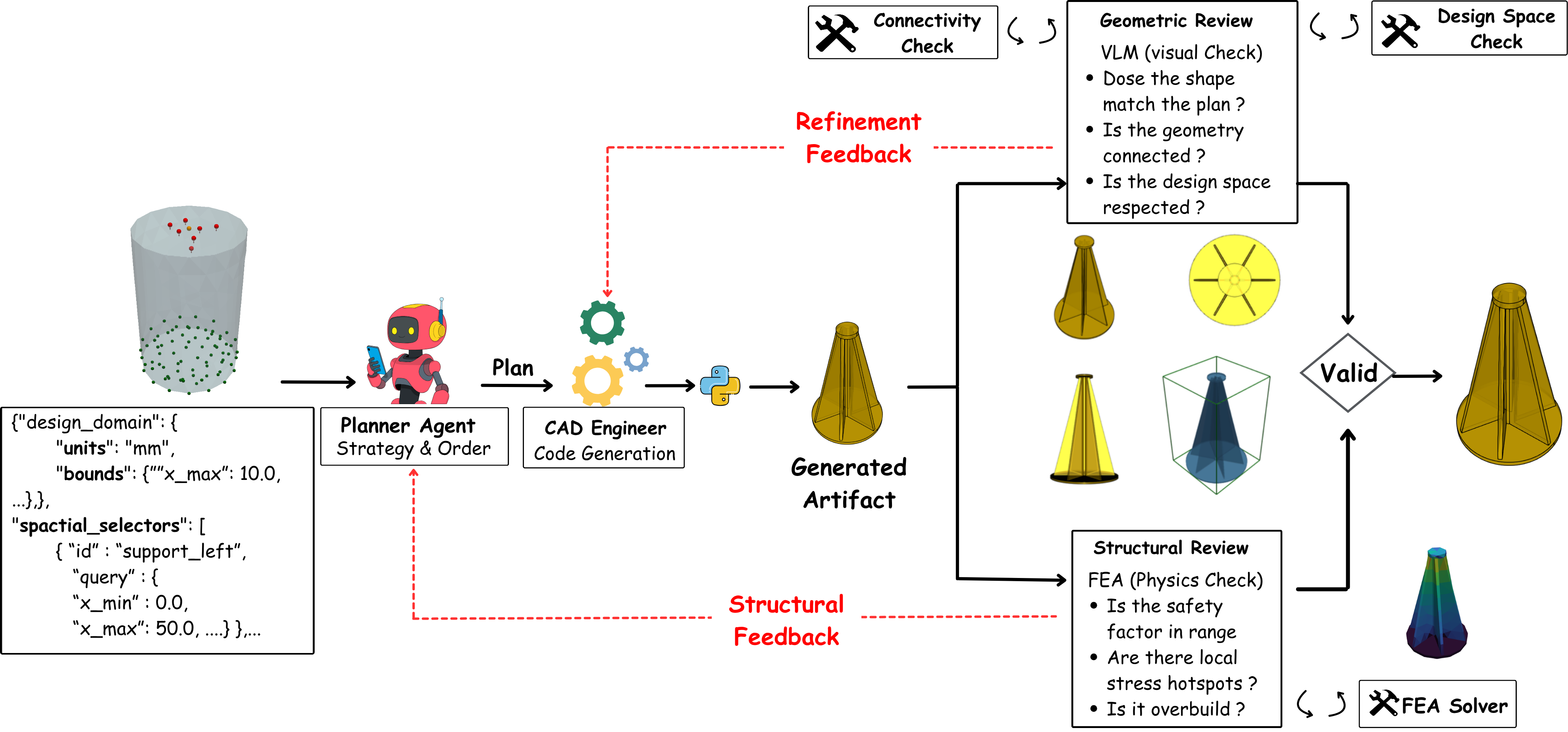}
\caption{Hybrid Agentic-Physical Architecture. The system processes structured load cases (left) through a multi-agent generation loop. The generated geometry (center) is subjected to parallel validation: a Vision-Language Model evaluates shape fidelity, while Finite Element Analysis rigorously verifies structural performance. Feedback from both streams guides the CAD Engineer in iterative refinement.}
\label{fig:architecture}
\end{figure*}

\section{System Architecture}

We propose a multi-agent framework built on LangGraph that coordinates specialized reasoning agents to enact an iterative ``Generate-Simulate-Refine'' engineering loop. This architecture formalizes the human design process as a graph of stateful nodes (current plan, CAD code, validator outputs) with directed cyclic execution. Feedback from downstream agents and physics-based tools directly informs upstream decision-making (see Figure \ref{fig:architecture}).

\subsection{Core Components}
The system is composed of four agents, orchestrated by a central control graph:

\textbf{Planner Agent.} This agent acts as the semantic bridge between user intent and geometric modeling. It decomposes load bearing requirements into a specific, step-by-step modeling plan. As input it receives the load case JSON and a visualization of the design space with boundary conditions and forces.

\textbf{CAD Engineer Agent.} The CAD Engineer consumes the structured plan and translates it into executable Python code using the CadQuery library. It ensures that the generated script is syntactically correct and topologically consistent. The CAD engineer agents receives feedback from the geometry reviewer to fix code and modelling errors.

\textbf{Geometry Reviewer Agent.} Receives as input the plan and renders of the CAD object from different angles. Acting as the critic, it checks if the geometry matches expectations, if the proposed part connects forces and constraints, is meshable, and enforces design space restrictions. It has physics-based tools for verifying connectivity and design space compliance. Its feedback is used to fix modelling errors.

\textbf{Structural Reviewer Agent.} Executes physics-based FEA to assess structural performance, evaluates whether the safety factor is within a specified target range, and identifies stress hotspots or over-engineering. Its feedback is used to refine the plan.

If any physics-based evaluation fails, the Orchestrator routes feedback to the Planner and CAD Engineer for refinement; agents may inspect but cannot modify deterministic evaluation results. A solution is considered valid if and only if all deterministic geometric and physics-based checks pass. Together with the source code and benchmark data, we will release the prompt templates.

\section{Experimental Setup}
We evaluate all models in an inference-only setting without any parameter training. A single disjoint load-case example is provided for in-context learning. The in-context example is excluded from the evaluation set and the model context is cleared after each run to prevent cross-instance information leakage. To assess the capabilities of the proposed agentic framework, we designed a benchmarking protocol that evaluates the system's ability to autonomously generate functional mechanical components from high-level specifications. We benchmark the performance of the architecture using four state-of-the-art VLMs as agents: Anthropic Claude 4.5 Sonnet, Anthropic Claude 4.5 Opus, Google Gemini 3 Pro, and Google Gemini 3 Flash. The task is design a component given the load case and achieve a safety factor within the target range while respecting the given design space. The agents are granted a maximum of 10 iterations per problem to converge on a valid solution. We perform three independent runs per load case and model to account for stochasticity in the generation process. 

\subsection{Evaluation Metrics}
In contrast to previous publications, we do not make use of geometry-based metrics, because in engineering design the functionality is of higher importance than geometric similarity to a reference design \cite{schulteFunctionalFeaturesDesign1993a}. Therefore, we introduce a new set of quantitative metrics to evaluate performance in reliability and design quality. These metrics are closely aligned with real-world engineering utility and are computed deterministically outside of VLMs.

\paragraph{Reliability.} 

\textbf{Geometry Generation Success Rate ($\text{R}_1$):} The percentage of generated CAD scripts that execute without geometric errors (e.g., extruding open sketches).
\textbf{Meshing Success Rate ($\text{R}_2$):} Fraction of geometries that can be successfully meshed. Modeling errors like non-manifold edges prevent meshing.
\textbf{FEA Success Rate ($\text{R}_3$):} The percentage of generated models that successfully pass the finite element solving stages. Errors like disconnected geometry cause failure.

\paragraph{Design quality.} \textbf{Safety Factor ($\text{DQ}_{\text{1}}$):} The minimum structural safety factor (yield strength / max stress). Higher values indicate robustness; excessive values indicate over-engineering.
\textbf{Structural Efficiency Ratio ($\text{DQ}_{\text{2}}$):} $SFR = \text{Safety Factor} / \text{Volume}$. Identifies designs optimizing the strength-weight trade-off.
\textbf{Number of Faces ($\text{DQ}_{\text{3}}$):} Number of unique faces in the B-Rep geometry, proxying geometric complexity. This metric captures the trade-off between expressiveness and complexity.
\textbf{Design Space Violation Rate ($\text{DQ}_4$):} Percentage of designs violating bounding volume constraints.
\textbf{Design Space Violation Magnitude ($\text{DQ}_5$):} The volume of material generated outside the allowable bounding box, normalized by design space volume.

\paragraph{Process efficiency.} We measure the average number of iterations required to reach a valid, target-compliant design \textbf{($\text{PE}_1$)}. If no valid design is found within 10 iterations, the run is counted as a failure.

We compare the different models and isolate the impact of physics-based feedback on reliability and design quality by performing statistical analyses.


\subsection{Implementation Details}
We used commercial cloud providers to run the LLMs, a machine with 48GB of VRAM to run the HXT meshing algorithm with \texttt{tet4} elements  and FEA, and the langgraph orchestration framework is executed on consumer hardware. We set the temperature to 0.5 to promote diverse results, allow for 4096 output tokens and disable thinking. On average a single design interation consumes 12,767 input tokens and 5,606 output tokens that, at the time of writing, costs approximately 0.023 USD (Gemini 3 Flash), 0.093 USD (Gemini 3 Pro), 0.204 USD (Claude Opus 4.5), or 0.122 USD (Claude Sonnet 4.5) in inference costs. The average wall-clock time of one iteration is 28.7$\pm$36.9 seconds.

\section{Results}

\begin{table}[t]
\centering

\begin{tabular}{lrrr}
\toprule
\textbf{Model} & \textbf{$\text{R}_1$} $\uparrow$ & \textbf{$\text{R}_2$} $\uparrow$ & \textbf{$\text{R}_3$} $\uparrow$ \\
\midrule
Claude Opus 4.5 & 96.4\% & \textbf{88.2\%} & 81.3\% \\
Claude Sonnet 4.5 & 97.5\% & 87.6\% & 84.5\% \\
Gemini 3 Flash & 97.0\% & 84.4\% & 85.1\% \\
Gemini 3 Pro & \textbf{97.5\%} & 76.3\% & \textbf{86.8\%} \\
\bottomrule
\end{tabular}%
\caption{Reliability Metrics: Execution Success ($R_1$) , Meshing Success ($R_2$) and Simulation Success ($R_3$).}
\label{tab:reliability}
\end{table}

\begin{table}[t]
\centering
\resizebox{\linewidth}{!}{%
\begin{tabular}{lrrrrr}
\toprule
\textbf{Model} & $\textbf{DQ}_1$ & $\textbf{DQ}_2$ $\uparrow$ & $\textbf{DQ}_3$ & $\textbf{DQ}_4$ $\downarrow$ & $\textbf{DQ}_5$ $\downarrow$ \\
\midrule
Claude Opus 4.5      & 4.97$\pm$1.6 & 37.2 & 85.0    & 27.7\%            & 0.3\%$\pm$1.0 \\
Claude Sonnet 4.5    & 4.96$\pm$1.5 & 14.5 & 99.7    & 85.7\%            & 39.6\%$\pm$64.5 \\
Gemini 3 Flash      & 4.04$\pm$1.5 & 22.7 & 63.1    & \textbf{8.7\%}    & \textbf{0.1\%}$\pm$0.3 \\
Gemini 3 Pro        & 4.18$\pm$1.5 & 24.3 & 120.1   & 32.3\%            & 12.7\%$\pm$46.1 \\
\bottomrule
\end{tabular}%
}
\caption{Design Quality Metrics: Safety Factor ($DQ_1$), Efficiency ($DQ_2$) , Geometric Complexity ($DQ_3$) and Constraint Compliance ($DQ_4, DQ_5$).}
\label{tab:design_quality}
\end{table}

\begin{table}[t]
\centering
\begin{tabular}{lrr}
\toprule
\textbf{Model} & \textbf{$\textbf{PE}_1$} Enabled $\downarrow$ & \textbf{$\textbf{PE}_1$} Disabled $\downarrow$ \\
\midrule
Claude Opus 4.5  & 6.0 & 10.4 \\
Claude Sonnet 4.5  & 4.5 & 13.0 \\
Gemini 3 Flash & 3.3 & 6.3 \\
Gemini 3 Pro & \textbf{2.5} & \textbf{4.0} \\
\bottomrule
\end{tabular}
\caption{Average Iterations to convergence ($PE_1$) with our agentic framework (Enabled) against single LLM (Disabled).}
\label{tab:process_efficiency_pe1}
\end{table}


The quantitative performance of the different models is summarized in Table \ref{tab:reliability} for reliability metrics and Table \ref{tab:design_quality} for design quality metrics. Additionally, Table \ref{tab:process_efficiency_pe1} presents the process efficiency in terms of iterations to convergence.
Inter-model comparisons revealed highly significant differences in code execution (H=163.04, p<0.01) and FEA success rates (H=46.24, p<0.01), with Gemini 3 Flash significantly outperforming all models (95.6\% execution, 84.4\% FEA; all pairwise p<0.01).

\section{Ablation}

In a series of ablation studies, we dissect the contributions of the physics-in-the-loop feedback mechanism and individual agent roles within the architecture. These experiments aim to isolate the impact of each component on overall system performance, providing insights into the efficacy of the hybrid design approach.

\subsection{Physics Feedback}

\begin{table}[t]
\centering
\resizebox{\linewidth}{!}{%
\begin{tabular}{lrrrrrr}
\toprule
& \multicolumn{3}{c}{\textbf{FEA Enabled}} & \multicolumn{3}{c}{\textbf{FEA Disabled}} \\
\cmidrule(lr){2-4} \cmidrule(lr){5-7}
\textbf{Model} & \textbf{Target} $\uparrow$ & \textbf{Under} $\downarrow$ & \textbf{Over} $\downarrow$ & \textbf{Target} $\uparrow$ & \textbf{Under} $\downarrow$ & \textbf{Over} $\downarrow$ \\
\midrule
Claude Opus 4.5  & \textbf{72.7\%} & 9.1\% & \textbf{18.2\%} & 14.3\% & 28.6\% & 57.1\% \\
Claude Sonnet 4.5  & 56.3\% & 25.0\% & 18.8\% & 25.0\% & 25.0\% & 50.0\% \\
Gemini 3 Flash & 68.2\% & \textbf{0.0\%} & 31.8\% & 20.0\% & \textbf{0.0\%} & 80.0\% \\
Gemini 3 Pro & 66.7\% & \textbf{0.0\%} & 33.3\% & \textbf{33.3\%} & 16.7\% & \textbf{50.0\%} \\
\bottomrule
\end{tabular}%
}
\caption{Comparison of CAD designs falling within the target safety factor range, underbuilt ($<2$) and overbuilt ($>5$) with FEA feedback enabled vs. disabled.}
\label{tab:fea_ablation}
\end{table}

We argue that embedding physics-based feedback from FEA into the design loop leads to a marked improvement in the functional validity of generated CAD models. To this end, we conduct an ablation study where we disable the physics-based tools (marked with \faWrench\ in Figure \ref{fig:architecture}) of the Reviewer Agents, allowing only feedback based on the visual appearance using VLMs. We observe the achieved safety factor ($DQ_1$) to assess the impact of physics feedback. We set a safety factor target range $[2, 5]$ consistent for general-purpose mechanical engineering \cite{budynas2020shigley}, and hypothesize that without physics feedback from FEA, designs will less consistently fall within this range (see Table \ref{tab:fea_ablation}). We performed a Fisher Exact Test \cite{fisher1922interpretation} comparing the success rates of achieving target safety factors between FEA-enabled (49/83 successes, 59.0\%) and FEA-disabled (6/27 successes, 22.2\%) conditions across all models, confirming statistical significance ($p = 0.0008$).

\subsection{Iterative Refinement}

\begin{figure}[h]
    \centering
    \begin{subfigure}{\linewidth}
        \centering
        \begin{minipage}{0.25\linewidth} \centering \includegraphics[width=\linewidth]{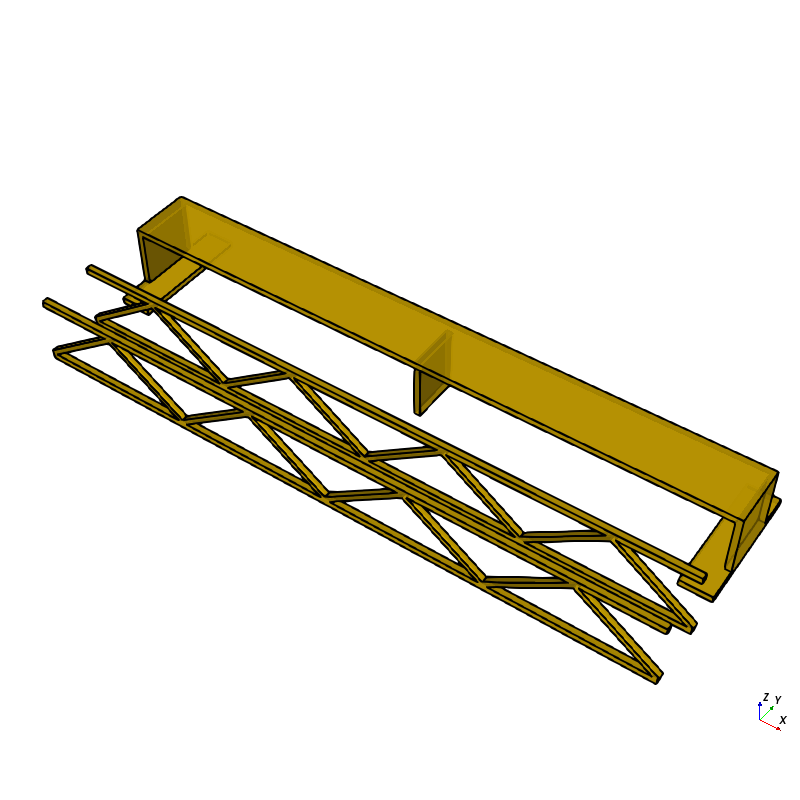} \\ \scriptsize Iter. 1 \end{minipage}
        \hfill
        \begin{minipage}{0.25\linewidth} \centering \includegraphics[width=\linewidth]{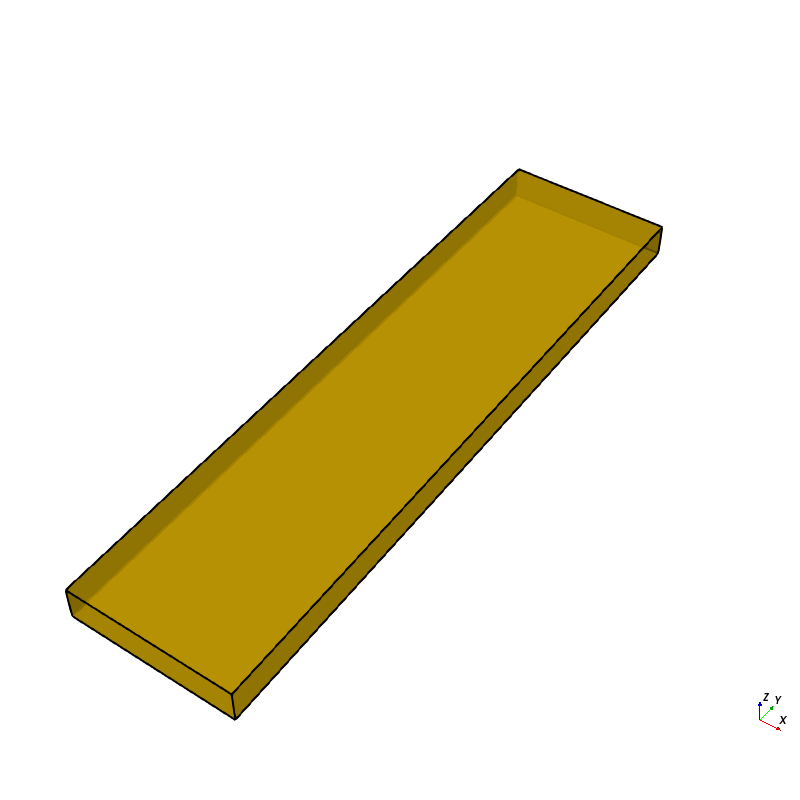} \\ \scriptsize Iter. 2 \end{minipage}
        \hfill
        \begin{minipage}{0.25\linewidth} \centering \includegraphics[width=\linewidth]{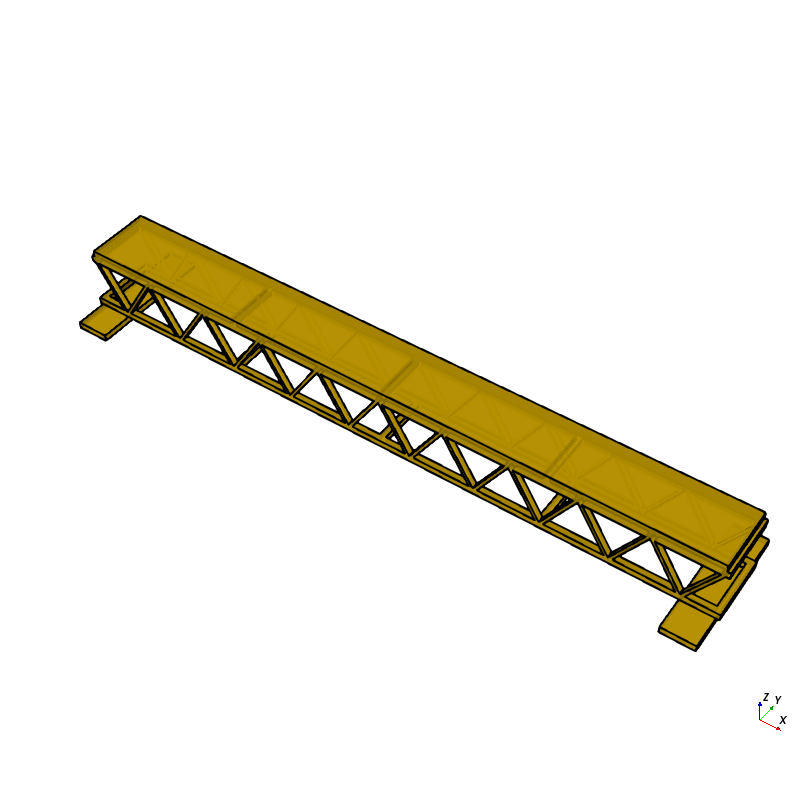} \\ \scriptsize Iter. 3 \end{minipage}
    \end{subfigure}

    \begin{subfigure}{\linewidth}
        \centering
        \begin{minipage}{0.19\linewidth} \centering \includegraphics[width=\linewidth]{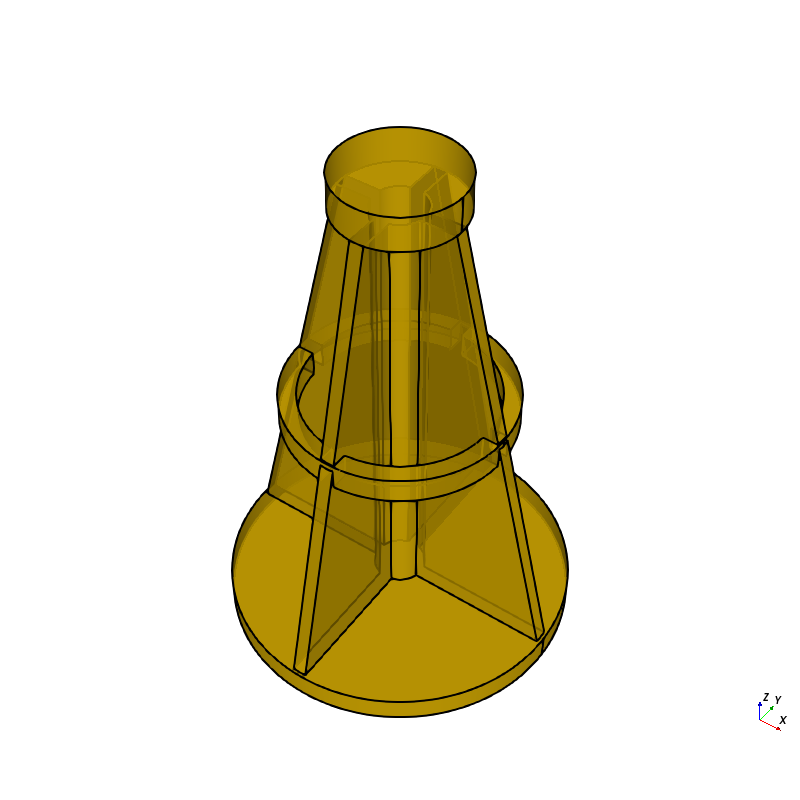} \\ \scriptsize Iter. 1 \end{minipage}
        \hfill
        \begin{minipage}{0.19\linewidth} \centering \includegraphics[width=\linewidth]{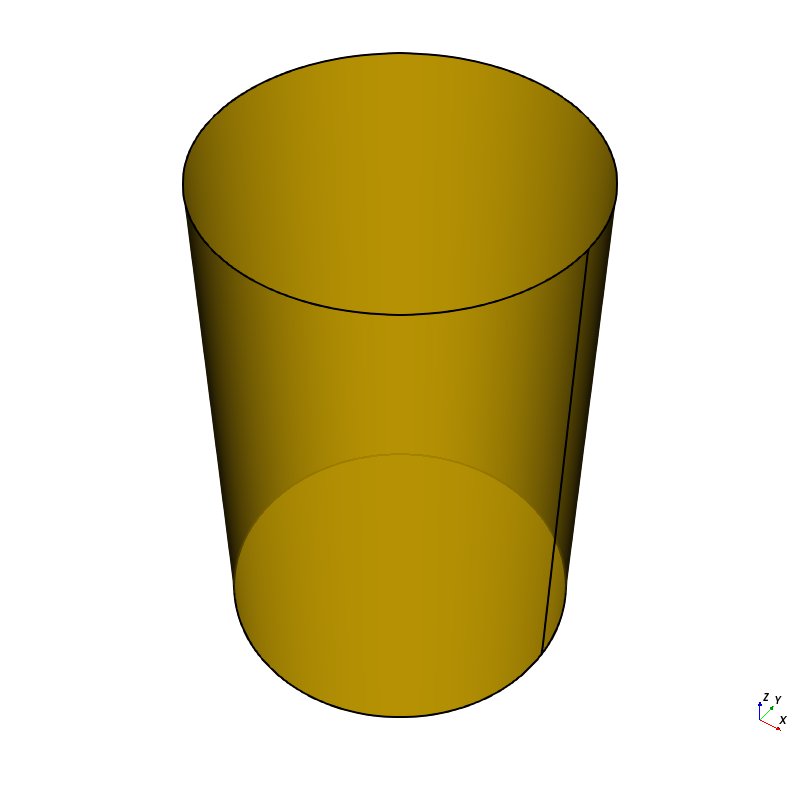} \\ \scriptsize Iter. 2 \end{minipage}
        \hfill
        \begin{minipage}{0.19\linewidth} \centering \includegraphics[width=\linewidth]{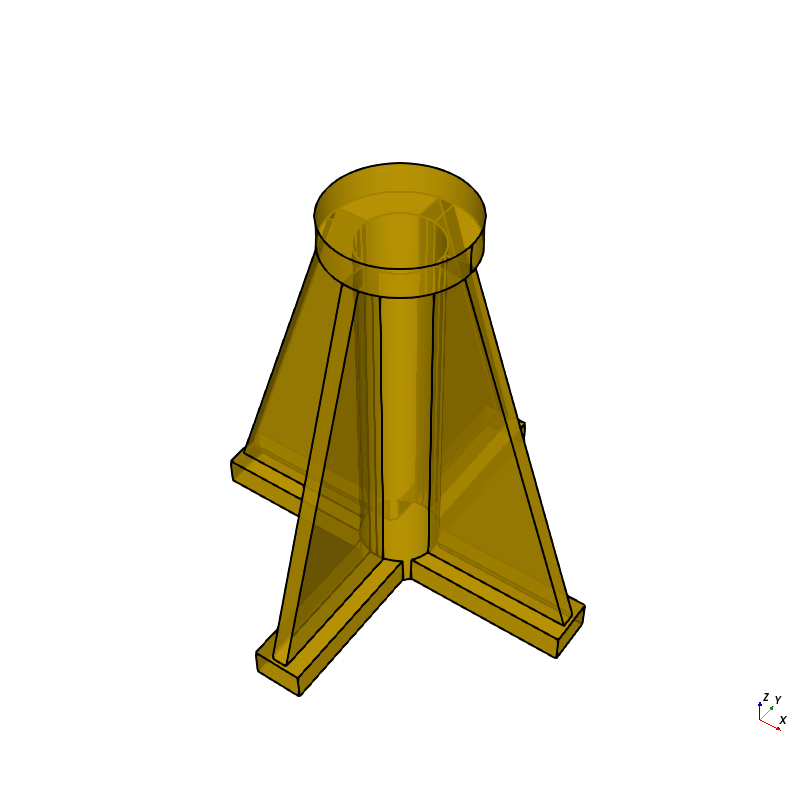} \\ \scriptsize Iter. 3 \end{minipage}
        \hfill
        \begin{minipage}{0.19\linewidth} \centering \includegraphics[width=\linewidth]{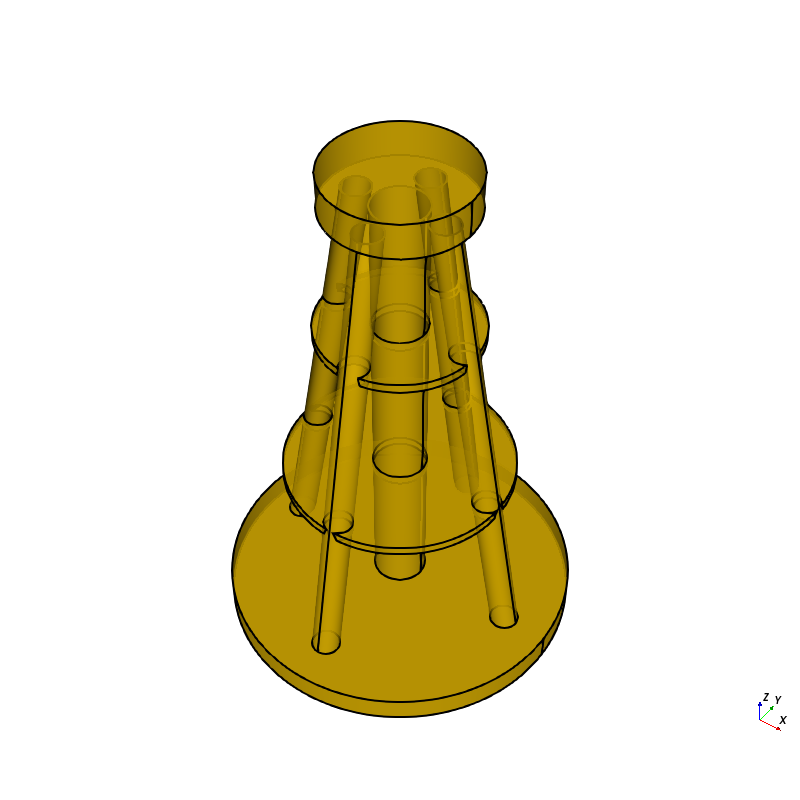} \\ \scriptsize Iter. 4 \end{minipage}
    \end{subfigure}

    \begin{subfigure}{\linewidth}
        \centering
        \begin{minipage}{0.25\linewidth} \centering \includegraphics[width=\linewidth]{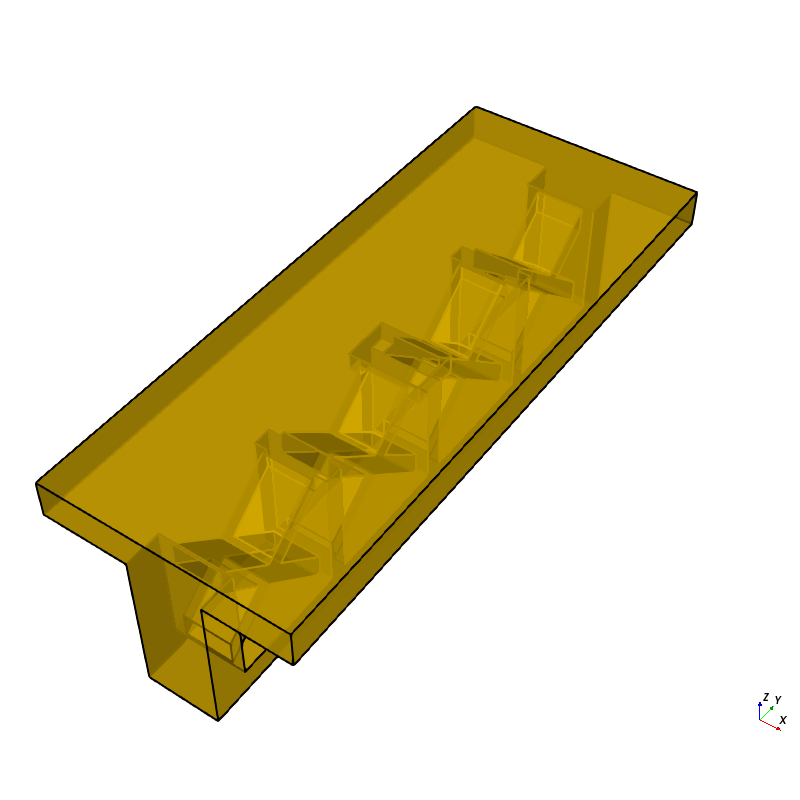} \\ \scriptsize Iter. 1 \end{minipage}
        \hfill
        \begin{minipage}{0.25\linewidth} \centering \includegraphics[width=\linewidth]{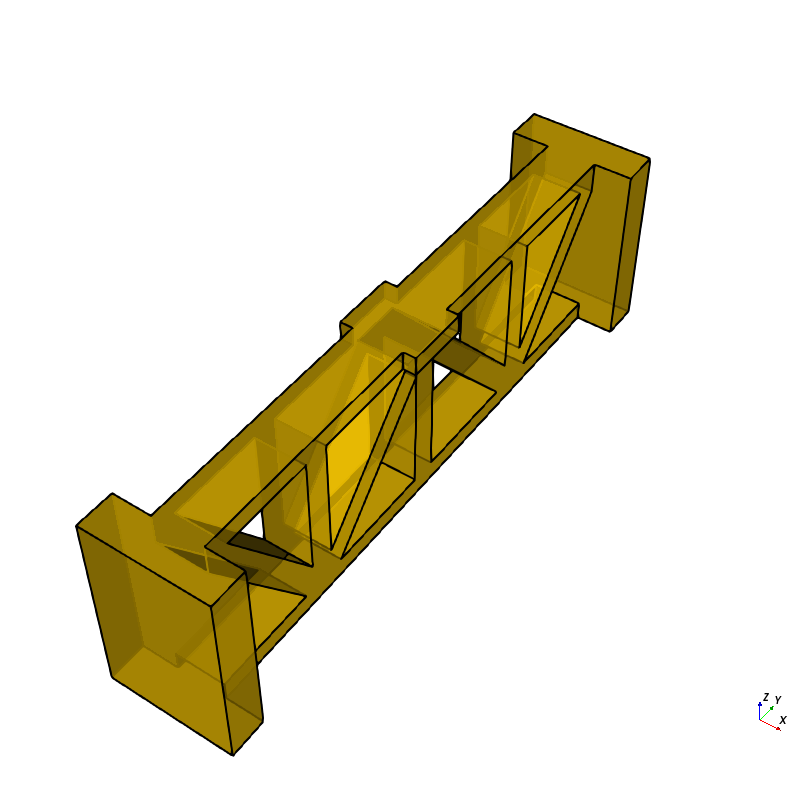} \\ \scriptsize Iter. 2 \end{minipage}
        \hfill
        \begin{minipage}{0.25\linewidth} \centering \includegraphics[width=\linewidth]{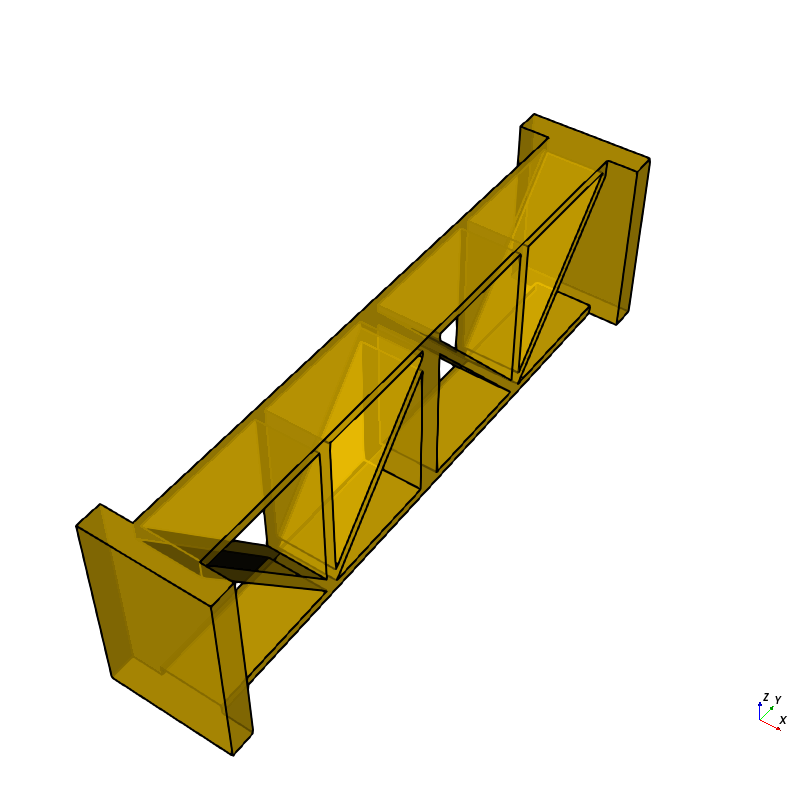} \\ \scriptsize Iter. 3 \end{minipage}
    \end{subfigure}
    
    \caption{Each row illustrates the iterative refinement of CAD models from initial agent-generated drafts to final physics-validated designs guided by FEA feedback.}
    \label{fig:iterative_refinement}
\end{figure}

\begin{figure}[t]
\centering
\begin{tikzpicture}
\pgfplotsset{
    every axis/.style={
        width=0.9\linewidth,
        height=4.8cm,
        grid style=dashed,
        ymajorgrids=true,
        xmin=1, xmax=15,
        xtick={1,2,3,4,5,6,7,8,9,10,11,12,13,14,15},
        label style={font=\small},
        tick label style={font=\footnotesize}
    }
}

\begin{axis}[
    name=top_plot,
    ymode=log,
    ylabel={Safety Factor},
    xticklabels={},
    extra y ticks={2,5},
    extra y tick style={grid=major, grid style={black, dashed, thick}},
]

\addplot[color=blue, mark=square] coordinates {
    (1,38.24)(2,18.84)(3,7.02)(4,24.90)(5,8.72)(6,12.89)(7,35.94)(8,4.68)(9,2.56)
};

\addplot[color=red, mark=triangle] coordinates {
    (1,10.08)(2,5.70)(3,6.15)(4,3.68)(5,2.89)(6,0.95)(7,0.66)(8,17.78)(9,75.13)(10,34.51)(11,1.24)(12,1.38)(13,3.29)
};

\addplot[color=green, mark=o] coordinates {
    (1,15.70)(2,9.94)(3,59.04)(4,30.02)(5,5.13)(6,4.82)(7,84.64)(8,22.73)(9,4.47)
};

\addplot[color=orange, mark=x] coordinates {
    (1,13.99)(2,12.94)(3,10.94)(4,3.93)(5,18.99)(6,16.57)(7,2.72)
};

\addplot[black, sharp plot, dashed, update limits=false] coordinates {(1,2) (15,2)};
\addplot[black, sharp plot, dashed, update limits=false] coordinates {(1,5) (15,5)};

\end{axis}

\begin{axis}[
    name=bottom_plot,
    at={(top_plot.south west)}, anchor=north west, yshift=-0.15cm,
    ylabel={Volume ($\text{cm}^3$)},
    xlabel={Iteration},
    legend columns=2,
    legend style={at={(0.5,-0.45)}, anchor=north, draw=none, font=\footnotesize},
]

\addplot[color=blue, mark=square] coordinates {
    (1,4222.02)(2,3373.90)(3,2357.32)(4,1820.68)(5,1434.68)(6,4740.45)(7,5254.84)(8,1574.09)(9,1249.60)
};
\addlegendentry{Claude Opus 4.5}

\addplot[color=red, mark=triangle] coordinates {
    (1,1950.87)(2,3599.60)(3,1587.27)(4,1241.20)(5,1597.92)(6,322.17)(7,506.13)(8,992.51)(9,1460.90)(10,6086.35)(11,425.09)(12,437.87)(13,355.58)(14,529.54)(15,426.86)
};
\addlegendentry{Claude Sonnet 4.5}

\addplot[color=green, mark=o] coordinates {
    (1,1857.72)(2,386.22)(3,1260.23)(4,487.48)(5,113.45)(6,518.56)(7,949.74)(8,544.64)(9,144.61)
};
\addlegendentry{Gemini 3 Flash}

\addplot[color=orange, mark=x] coordinates {
    (1,3312.96)(2,1816.59)(3,2124.78)(4,1145.58)(5,3243.18)(7,5233.93)(8,2451.11)
};
\addlegendentry{Gemini 3 Pro}

\end{axis}
\end{tikzpicture}
\caption{Safety Factor (target range 2--5 dashed, top) and volume (bottom) convergence. Physics feedback guides agents to converge on valid safety factors while progressively reducing volume.}
\label{fig:metrics_convergence}
\end{figure}
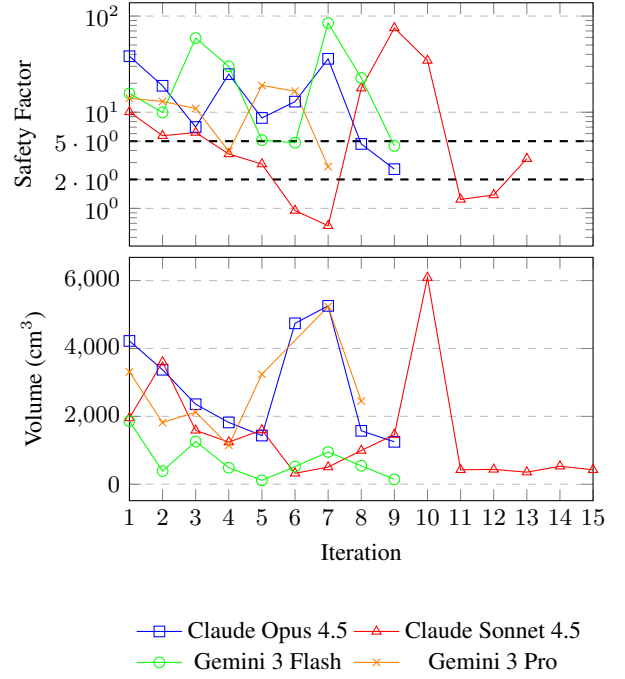

To isolate the effect of iterative refinement driven by physics feedback, we compare the performance metrics at each iteration of the design loop. Figure~\ref{fig:metrics_convergence} illustrates the evolution of the safety factor across iterations. The results show that the safety factor initially exceeds the target range but gradually improves with each iteration, converging to an acceptable level after several rounds of refinement. This underscores the efficacy of the physics-in-the-loop feedback mechanism in guiding the design process towards structurally sound~solutions.

\subsection{Multi-Agent}

Next, we aim to isolate the impact of a multi-agent setup vs. a single LLM and remove the planning and review agents. Instead we formulate the task with a single CAD Engineer agent with access to the physics-based tools that directly generates CAD code from the load case description and refines output from raw FEA results. We observe the average safety factor ($DQ_1$) and convergence ($PE_1$) for randomly sampled inputs in Table \ref{tab:process_efficiency_pe1} to assess if the planner reduces the load of the engineer, accelerating convergence and if the reviewer's structured feedback improves safety factor. We performed a Welch t-test comparing iterations required with planning enabled ($n=80$, $M=4.44$, $SD=5.36$) versus disabled ($n=80$, $M=10.77$, $SD=10.11$), confirming a significant increase in required iterations without planning ($t=-3.17$, $df=13.39$, $p=0.0071$, Cohen's $d=-1.03$).

\section{Discussion}

We observe that our system is capable of correctly interpreting given load cases and generate complex CAD designs. Visual analysis of the generated CAD code indicates that Claude Sonnet 4.5 frequently produces overly complex geometries that are difficult to mesh and simulate, resulting in lower FEA success rates and high design space violations. In contrast, Gemini 3 Pro generates more efficient designs that balance structural complexity with manufacturability. Notably, the smaller Gemini 3 Flash outperforms Claude Sonnet 4.5 despite its reduced model size. This reveals a "Goldilocks effect" where the mid-tier Gemini 3 Flash excels, likely because it generates simpler, more practical geometries that are easier to mesh and simulate, whereas larger LLMs over-engineer, resulting in modelling errors.

Inspecting the average safety factors in Table \ref{tab:design_quality} of the approved designs, we observe that all models tend to over-engineer, producing safety factors close to the upper limit of the target range (5.0). Opus 4.5 hit the iteration limit most frequently (16.3\%), followed by Sonnet 4.5 (12.8\%).

\subsection{Planning the CAD Design}
We see that the Gemini models deteriorate less from the removal of the Planner Agent (see results in Table \ref{tab:process_efficiency_pe1}). This suggests that these models possess sufficient reasoning capabilities to decompose the design task internally. However, the Planner Agent still provides value by structuring the CAD Engineer's workflow, leading to more consistent code generation. Without the Planner Agent Claude Sonnet 4.5 and Claude Opus 4.5 exhibit larger degradation, indicating that explicit task decomposition is more important for their~success.

\subsection{Physics-in-the-Loop}

The ablation studies indicate that embedding physics feedback in the design loop enables the system to more effectively correct over- and under-engineered solutions. The FEA validation provides concrete, quantitative performance measures that help guide the CAD Engineer in making targeted design adjustments. This feedback loop enables the system to iteratively refine designs towards optimal structural performance, reducing reliance on geometric heuristics alone. However, with more iterations, the results tend to become more unstable. Future work could focus on detecting stopping criteria when a CAD design becomes unrecoverable.

\subsection{Failure Type Analysis}

\begin{figure}[ht]
\centering
\begin{tikzpicture}
\begin{axis}[
    ybar,
    width=0.95\linewidth,
    height=4.5cm,
    ymin=0, ymax=50,
    ylabel={Count},
    xtick={1,2,3,4,5},
    xticklabels={Design Space, Connectivity, FEA, Load Area, Fix Area},
    tick label style={font=\scriptsize, align=center},
    legend style={at={(0.5,-0.3)}, anchor=north, legend columns=4, draw=none, font=\scriptsize},
    enlarge x limits=0.15,
    bar width=4pt,
    grid=major,
    major grid style={dotted, gray!50},
    cycle list={
        {draw=blue, fill=blue!30},
        {draw=red, fill=red!30},
        {draw=green!60!black, fill=green!30},
        {draw=orange, fill=orange!30}
    }
]
\addplot coordinates {(1,33) (2,31) (3,2) (4,1) (5,0)};
\addplot coordinates {(1,41) (2,42) (3,18) (4,4) (5,4)};
\addplot coordinates {(1,32) (2,28) (3,2) (4,3) (5,0)};
\addplot coordinates {(1,16) (2,17) (3,1) (4,2) (5,0)};

\legend{Opus, Sonnet, Flash, Pro}
\end{axis}
\end{tikzpicture}
\caption{Distribution of failure types by model that can occur due to lacking comprehension of the design task or modelling errors.}
\label{fig:failure_modes_chart}
\end{figure}
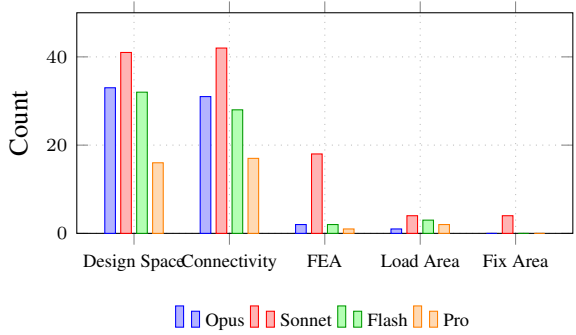

Designs most commonly fail due to design space violations and disconnected parts. The distribution is consistent across all tested LLMs with Claude Sonnet-4.5 performing worst across of all failure cases. We attribute high design space violations and connectivity errors to the limited spatial reasoning for load paths of LLMs. FEA-related failures are relatively rare, suggesting that when geometries pass meshing, they are generally well-posed for simulation. Rarely it happens that the area where loads or fixed supports are applied is not filled with material. Figure \ref{fig:failure_modes_chart} summarizes the failure type distribution.

\subsection{Comparison to Other Methods}
\begin{figure}
    \centering
    \begin{subfigure}{0.25\linewidth}
        \centering
        \includegraphics[width=\linewidth]{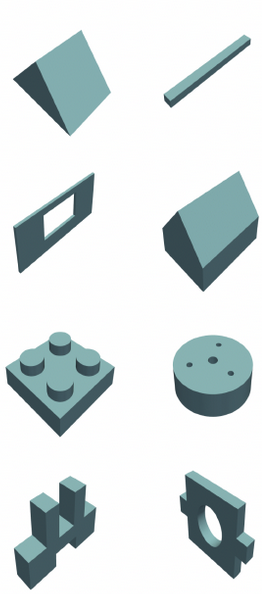}
        \caption{}
        \label{fig:cad_code_verify}
    \end{subfigure}
    \hfill
    \begin{subfigure}{0.32\linewidth}
        \centering
        \includegraphics[width=\linewidth]{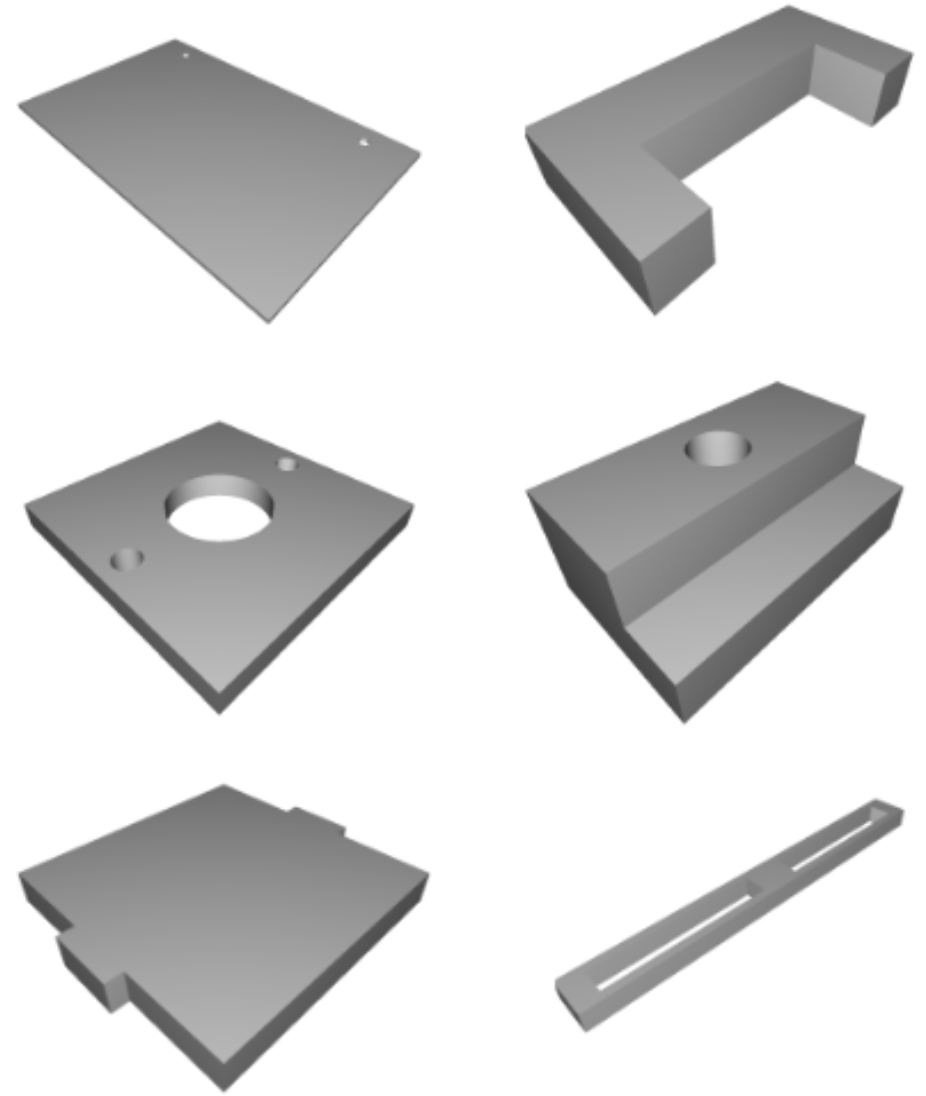}
        \caption{}
        \label{fig:evocad}
    \end{subfigure}
    \hfill
    \begin{subfigure}{0.32\linewidth}
        \centering
        \includegraphics[width=\linewidth]{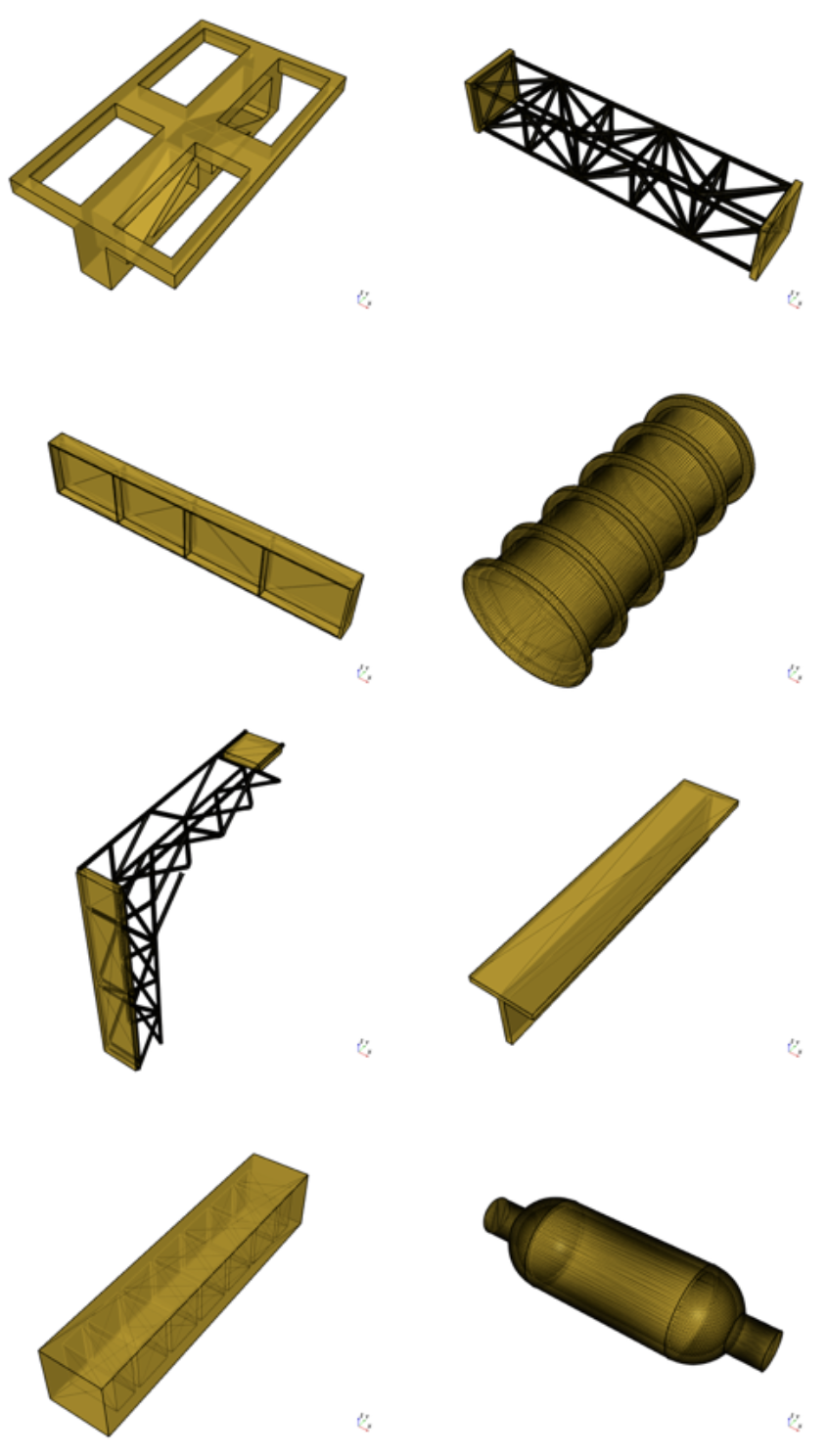}
        \caption{}
        \label{fig:our_method}
    \end{subfigure}
    
    \vspace{0.3cm}
    
    \begin{subfigure}{0.24\linewidth}
        \centering
        \includegraphics[width=\linewidth]{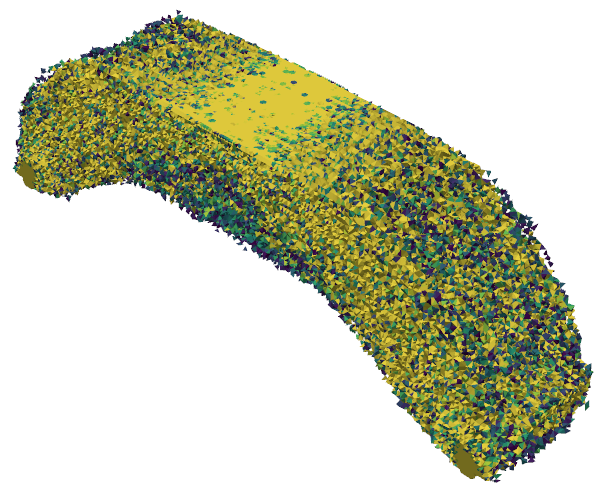}
        \caption{ }
        \label{fig:to_arch_bridge}
    \end{subfigure}
    \begin{subfigure}{0.24\linewidth}
        \centering
        \includegraphics[width=\linewidth]{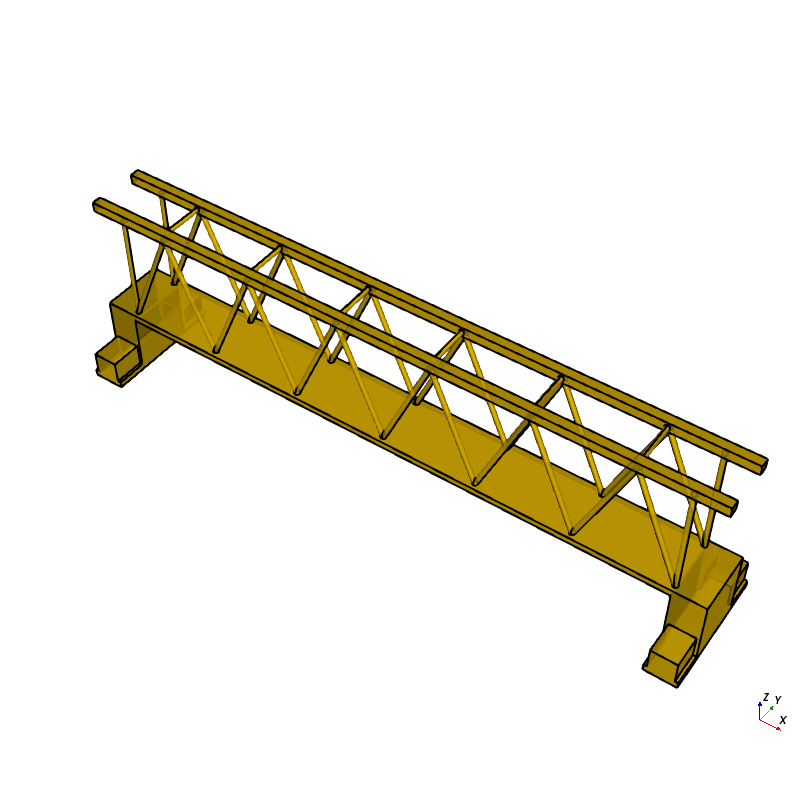}
        \caption{}
        \label{fig:gemini_pro}
    \end{subfigure}
    \begin{subfigure}{0.24\linewidth}
        \centering
        \includegraphics[width=\linewidth]{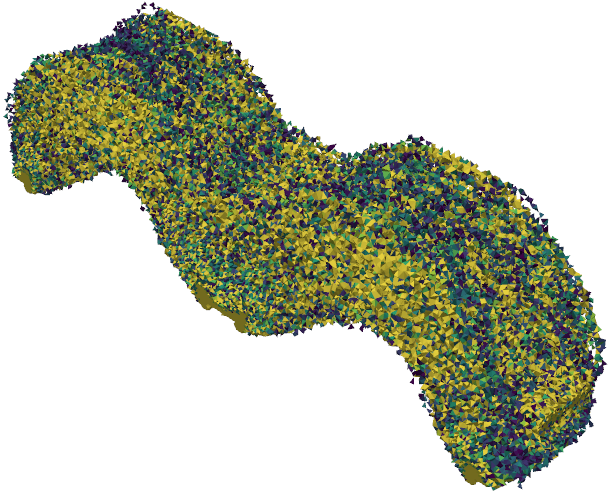}
        \caption{}
        \label{fig:to_double_arch_bridge}
    \end{subfigure}
    \begin{subfigure}{0.24\linewidth}
        \centering
        \includegraphics[width=\linewidth]{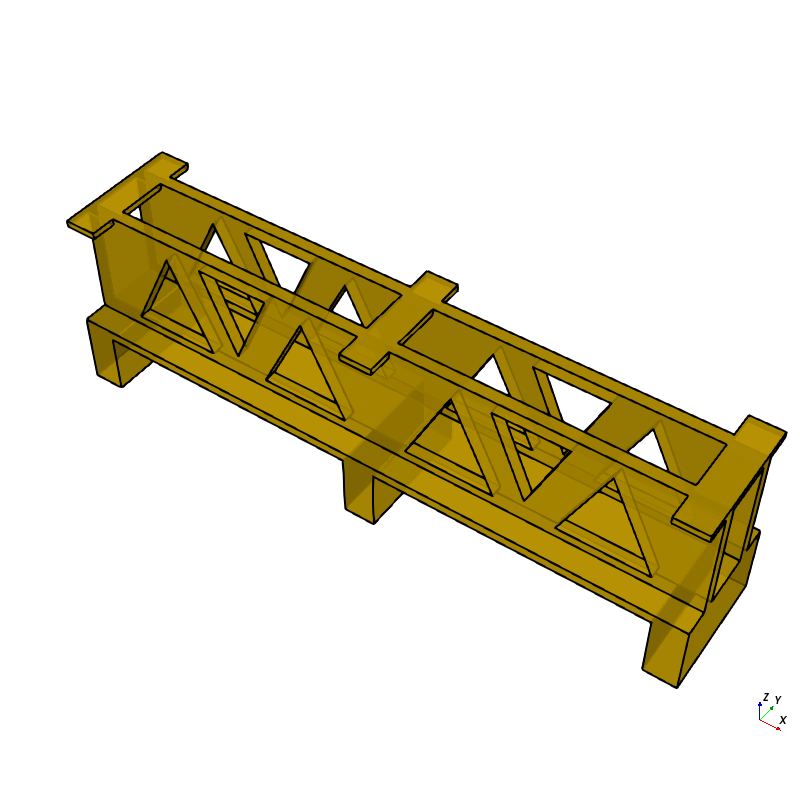}
        \caption{}
        \label{fig:our_double_arch_bridge}
    \end{subfigure}
    \caption{Qualitative comparison showing the fidelity of (a) CadCodeVerify \protect\cite{alrashedyGeneratingCADCode2025},  (b) EvoCAD \protect\cite{preintner2025evocadevolutionarycadcode}, (c) our method. We compare outputs of topology optimization (d, f), and our method (e, g) the same load case with our method.}
    \label{fig:qualitative_comparison}
\end{figure}

Since our method is novel in its problem formulation, there are no directly comparable prior works. However, compared to purely generative CAD systems that lack physics validation, our approach demonstrates  3.4\% higher compile rate ($IR_1$) compared to \citeauthor{alrashedyGeneratingCADCode2025}~\shortcite{alrashedyGeneratingCADCode2025}. We can also compare the complexity of generated geometries: our method achieves an average of 83.1 faces ($DQ_3$) compared to 24.2 faces present in methods based on the DeepCAD dataset \cite{wuDeepCADDeepGenerative2021} such as the method by \citeauthor{alrashedyGeneratingCADCode2025} and \citeauthor{preintner2025evocadevolutionarycadcode}. A qualitative comparison is shown in Figure~\ref{fig:qualitative_comparison}.

\subsection{Comparison to Topology Optimization}
Topology optimization (TO) methods \cite{BendsoeSigmund2013} solves the same problem. We compare the execution time of our method on our dataset with a PyTorch-based TO implementation\footnote{https://github.com/meyer-nils/torch-fem} on a machine with 48GB VRAM, with 10 iterations and same meshing parameters and found that TO takes on average 18.6$\pm$1.2 seconds to converge whereas our method requires 28.7$\pm$36.9 seconds. However, our approach has advantage that it produces editable CAD models directly with higher fidelity (example shown in Figure \ref{fig:qualitative_comparison}) while TO produce voxel or mesh-based representations that require additional post-processing like manually tracing design in CAD software.
\section{Conclusion}

Extensions to dynamic or thermal scenarios or optimization for further criteria, such as sustainability, are left for future research. This can include integrating material selection, manufacturability assessments, and cost-analysis into the design loop. Architecturally, the current prompt-driven incorporation of physics feedback could be replaced with reinforcement learning-based optimization.

We present a hybrid agentic-physical system that combines LLM-based agents with mature engineering tools. The proposed approach generates CAD designs based on load bearing requirements that are structurally grounded and more geometrically complex than those generated by geometry optimized methods, by embedding physics-based validation directly into the agents' decision-making loop. Our results suggest that combining agentic reasoning with knowledge-based engineering tools offers a practical foundation for trustworthy AI-assisted systems in engineering design.

\bibliographystyle{named}
\bibliography{ijcai26}

\end{document}